\pdfoutput=1
\documentclass[11pt]{article}
\usepackage[final]{acl}
\usepackage{times}
\usepackage{latexsym}
\usepackage{microtype}
\usepackage{graphicx}
\usepackage{amsmath,amssymb,mathtools}
\usepackage{booktabs}
\usepackage{multirow}
\usepackage{tabularx}
\usepackage{array}
\usepackage{enumitem}
\usepackage{tikz}
\usetikzlibrary{arrows.meta,positioning,fit,calc,backgrounds,shapes.geometric,decorations.pathreplacing,matrix}
\usepackage{pgfplots}
\pgfplotsset{compat=1.18}
\usepgfplotslibrary{polar}
\usepackage{url}

\makeatletter
\AtBeginDocument{%
  \IfFormatAtLeastTF{2025-06-01}{%
    \RemoveFromHook{build/column/before}[lineno]%
    \def\@LN@makecol{%
      \@LN@orig@makecol
      \setbox\@outputbox\vbox{%
        \boxmaxdepth\@maxdepth
        \protected@write\@auxout{}{%
          \string\def\string\@LN@column{\if@firstcolumn1\else2\fi}}%
        \box\@outputbox}}%
  }{}%
}
\makeatother

\title{DINO Eats Geometry: Self-Supervised Sign Representations\\
that Absorb Articulated Hand Geometry}

\author{Junyi Hu \quad Zhewen He \quad Haomian Huang \quad Zhenhua Li \\
  \textbf{Shuaihang Yuan} \quad \textbf{Zhifei Li} \quad \textbf{Yi Fang}\thanks{\ \ Corresponding author.} \\
  New York University Abu Dhabi \\
  \texttt{jh10472@nyu.edu}}
\newcommand{\sdeg}{\textsc{SignDEG}}
\newcommand{\ours}{\sdeg}
\newcommand{\signdinobaseline}{\textsc{SignDINO}}
\newcommand{\shubert}{\textsc{SHuBERT}}

\newcommand{\dinov}{\textsc{DINOv2}}
\newcommand{\dinovthree}{\textsc{DINOv3}}

\newcommand{\loss}{\mathcal{L}}

\newcolumntype{Y}{>{\raggedright\arraybackslash}X}
\newcolumntype{C}{>{\centering\arraybackslash}X}

\definecolor{cvideo}{HTML}{DCEBFF}
\definecolor{cregion}{HTML}{E6E6E6}
\definecolor{ctube}{HTML}{E5F7E5}
\definecolor{cstudent}{HTML}{E7DFFF}
\definecolor{cteacher}{HTML}{FFF2B2}
\definecolor{closs}{HTML}{EEEEEE}
\definecolor{cproto}{HTML}{DDF7F4}
\definecolor{cgram}{HTML}{F3D9FA}
\definecolor{cdarkblue}{HTML}{1F4E79}
\definecolor{cdarkgreen}{HTML}{2E7D32}
\definecolor{cdarkred}{HTML}{000000}
\definecolor{cdarkpurple}{HTML}{5E35B1}
\definecolor{cdarkorange}{HTML}{000000}
\definecolor{cgray}{HTML}{F2F2F2}

\begin{document}
\maketitle

\begin{figure*}[t]
\centering
\includegraphics[width=\textwidth]{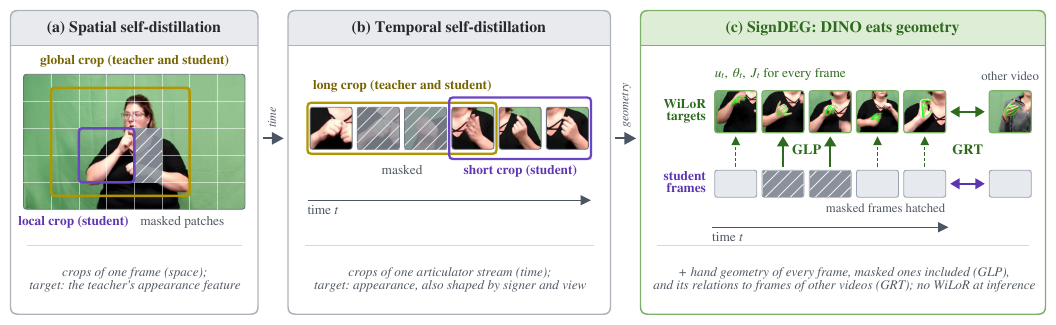}
\caption{From spatial to temporal self-distillation to \sdeg. \emph{(a)} Image DINO: a student matches an EMA teacher across global and local crops of one frame. \emph{(b)} Temporal self-distillation plays the same game on long and short temporal crops of one tracked articulator stream, with some frames masked; in both, the target is the teacher's own appearance feature. \emph{(c)} During pre-training, \sdeg\ also asks the student to predict, at every frame including masked ones, the geometry that a frozen hand reconstruction model (WiLoR) assigns to the crop (GLP), and to transfer geometric relations between frames of different videos (GRT). WiLoR is removed at inference.}
\label{fig:insight}
\end{figure*}
\vspace{0.4em}

\begin{abstract}
Handshape and orientation are geometric components of sign language, yet many sign encoders learn from appearance alone. We introduce \sdeg\ (\emph{DINO Eats Geometry}), a temporal sign encoder that uses articulated hand geometry as a pre-training target and requires only video at inference. Frozen \dinovthree\ features of tracked hands and faces feed per-stream temporal Transformers. Geometric latent prediction (GLP) supervises an intermediate block with WiLoR latents, MANO articulation, and 3D joints, including at masked frames. Geometric relation transfer (GRT) matches cross-video frame-similarity distributions at the final block. Both objectives are learned jointly with temporal self-distillation; WiLoR and the geometry heads are then discarded. Under the \shubert\ public-data evaluation protocol, \sdeg\ reaches 17.9/24.5 BLEU-4 on How2Sign/OpenASL, compared with 16.2/23.2 for \shubert. Rank-1 adaptation improves ASL Citizen and Sem-Lex recognition by 0.02 Recall@1 each, and phonological classification improves across 16 attributes. Removing GRT while retaining GLP reduces How2Sign BLEU-4 by 1.4 and ASL Citizen Recall@1 by 0.034. The results support transferring articulated structure into an appearance encoder without running a hand reconstruction model at deployment.
\end{abstract}

\section{Introduction}
\label{sec:intro}

Gloss-free sign language translation (SLT) maps signing video to spoken-language text without intermediate gloss annotations~\citep{lin2023glofevn,wong2024sign2gpt,li2025unisign}. Scarce parallel data motivate self-supervised sign encoders~\citep{hu2021signbert,zhao2024masa,rust2024ssvpslt,gueuwou2025shubert,gan2026signdino}. Sign phonology decomposes a sign into handshape, palm orientation, body-relative location, and movement, together with non-manual markers~\citep{stokoe1960sign,brentari1998prosodic}. Handshape and orientation describe an articulated state; movement organises those states over time. Learning a useful sign representation therefore requires both fine articulatory structure and temporal context.

Appearance encoders supply strong visual features but do not explicitly supervise this articulated state. Whole-frame DINO adaptation~\citep{gan2026signdino} and temporal self-distillation over tracked appearance crops learn from image features whose similarity can reflect viewpoint or signer appearance as well as handshape. Systems that take pose as input~\citep{hu2023signbertplus,li2025unisign} make articulation explicit, but require a pose estimator at inference. A complementary approach is to use geometry as a learning target: the encoder can acquire articulatory structure during pre-training while retaining an appearance-only inference pathway.

We propose \sdeg\ (Figure~\ref{fig:insight}). A frozen \dinovthree\ maps tracked hand and face crops to appearance features, and a temporal Transformer encodes each stream. During pre-training, a frozen hand reconstruction model supplies geometric targets to the hand streams. GLP predicts its latent, MANO articulation, and 3D joints from an intermediate block; masked frames require prediction from neighbouring motion. GRT aligns the final block's cross-video similarity distributions with those of the geometric targets. This division places per-frame articulation within the temporal encoder and geometric relations at its output. Both losses are active from the first update; the hand model is removed at inference. Using an auxiliary foundation-model space for training has proved useful in representation alignment and geometric generation~\citep{yu2025repa,jang2026gld}; here the target is the articulated hand, and the downstream objective is sign representation learning.

Our contributions are: (i)~a sign encoder that distils articulated hand geometry into temporal appearance features without geometry estimation at inference; (ii)~masked geometric prediction at intermediate depth and cross-video geometric relation transfer at the final block, with reliability-weighted targets; and (iii)~evaluation across translation, recognition, fingerspelling, and phonological categories. Frozen features improve How2Sign translation over raw \dinovthree\ crops; rank-1 adaptation improves ASL Citizen recognition over \shubert. Matched geometry controls and label-based retrieval identify the structure contributed by the geometric objectives.

\section{Related Work}
\label{sec:related}

\paragraph{Self-supervised sign representations.}
Sign SSL has used hand-pose reconstruction~\citep{hu2021signbert,hu2023signbertplus}, masked video modelling~\citep{zhao2024masa,rust2024ssvpslt}, sign--text alignment~\citep{jiang2024signclip}, and prediction of clustered units over hand, face, and body streams~\citep{gueuwou2025shubert}. Multi-stream recognition~\citep{chen2022twostream,yin2020stmc} and graph-based translation~\citep{gan2025mixsigngraph} model complementary spatial and temporal cues. The CVPR \signdinobaseline~\citep{gan2026signdino} adapts frame-level DINO using whole-frame and articulator-focused views. \sdeg\ instead applies self-distillation along the time axis of tracked articulator streams and supervises this pathway with fixed hand-geometry targets. Geometry is a pre-training target rather than a downstream input; source signing videos need no gloss, translation, or geometry annotations for this stage.

\paragraph{Foundation-model features as learning targets.}
Representation alignment~\citep{yu2025repa} and geometric latent diffusion~\citep{jang2026gld,lin2025da3,wang2025vggt} use foundation-model features to expose structure useful for another learning objective. DINO Eats CLIP~\citep{he2026dec} uses CLIP-guided virtual features for open-set 3D retrieval. Relational distillation transfers pairwise structure between feature spaces~\citep{park2019rkd}. WiLoR and HaMeR recover MANO articulation and 3D joints from hand crops~\citep{romero2017mano,potamias2025wilor,pavlakos2024hamer}. Our contribution combines per-frame articulated targets with masked temporal prediction and cross-video relation transfer for sign language. WiLoR contributes a supervised geometric prior learned outside the signing corpus; pre-training the sign encoder itself uses no new manual annotations.

\section{\sdeg}
\label{sec:method}

\begin{figure*}[t]
\centering
\includegraphics[width=\textwidth]{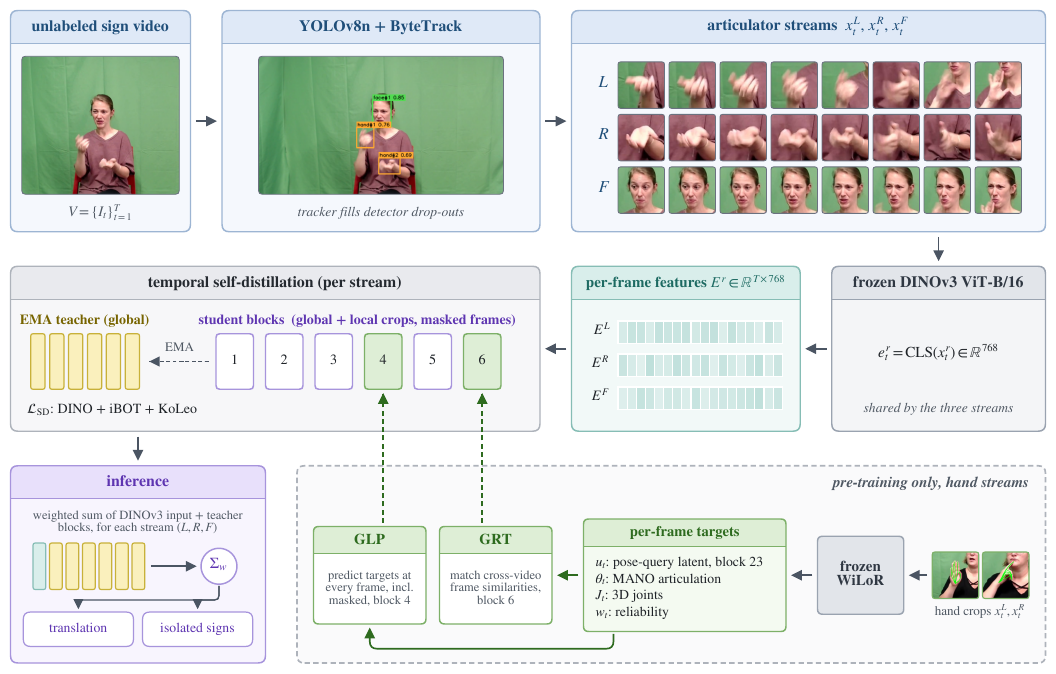}
\caption{Overview of \sdeg. \emph{Top:} a detector and tracker extract left-hand, right-hand, and face crop streams from each video. \emph{Middle (right to left):} a frozen \dinovthree\ embeds every crop, and a temporal student and its EMA teacher are trained per stream by temporal self-distillation. \emph{Bottom right:} during pre-training only, a frozen hand reconstruction model supplies per-frame geometric targets for the hand streams; the student predicts them from block 4, including at masked frames (GLP), and aligns its cross-video frame similarities at block 6 with geometric similarities (GRT). \emph{Bottom left:} at inference, the EMA teachers run on \dinovthree\ features alone.}
\label{fig:overview}
\vspace{-0.3em}
\end{figure*}
\vspace{0.4em}

\sdeg\ uses a temporal encoder per articulator stream, with objectives assigned to different depths (Figure~\ref{fig:overview}). Its input is appearance: a frozen \dinovthree\ embeds tracked crops (\S\ref{sec:method:upstream}). Self-distillation between a student and an EMA teacher trains its output to describe how the stream evolves (\S\ref{sec:method:ssl}). During pre-training, a frozen hand foundation model supplies per-frame geometric targets (\S\ref{sec:method:targets}); an intermediate block learns to predict them at every frame (\S\ref{sec:method:glp}), and the last block learns cross-video frame relations in geometric space (\S\ref{sec:method:grt}). Inference retains only the appearance pathway and the EMA teacher (\S\ref{sec:method:obj}). Throughout, an \emph{articulator crop} is the image region of one hand or the face in one frame, and a \emph{temporal crop} is a contiguous run of frames of one stream.

\subsection{Articulator streams}
\label{sec:method:upstream}

The hands and face occupy small regions of the frame and move independently, so each is tracked as a separate stream. A hand and face detector (YOLOv8n) runs on every frame, and ByteTrack~\citep{zhang2022bytetrack} associates hand boxes with two persistent identities and fills short detector gaps with its motion model. Each video yields three streams $r\in\{\mathrm{L},\mathrm{R},\mathrm{F}\}$ (left hand, right hand, face); $x^{r}_t$ is the square articulator crop of stream $r$ at frame $t$, and $m^{r}_t\in\{0,1\}$ marks whether the tracker produced a box. A frozen \dinovthree\ ViT-B/16~\citep{simeoni2025dinov3} maps each crop to its class token:
\begin{equation}
e^{r}_t = \mathrm{CLS}\big(x^{r}_t\big) \in \mathbb{R}^{768}.
\label{eq:dinov3_embed}
\end{equation}
These frame embeddings are cached once per video. Each stream has its own temporal Transformer, trained independently; we drop the index $r$ below. Frame rate, crop sizes, and all other hyperparameters are listed in Appendix~\ref{app:sdeg_hparams}.

\subsection{Temporal-axis self-distillation}
\label{sec:method:ssl}

\begin{figure*}[t]
\centering
\includegraphics[width=\textwidth]{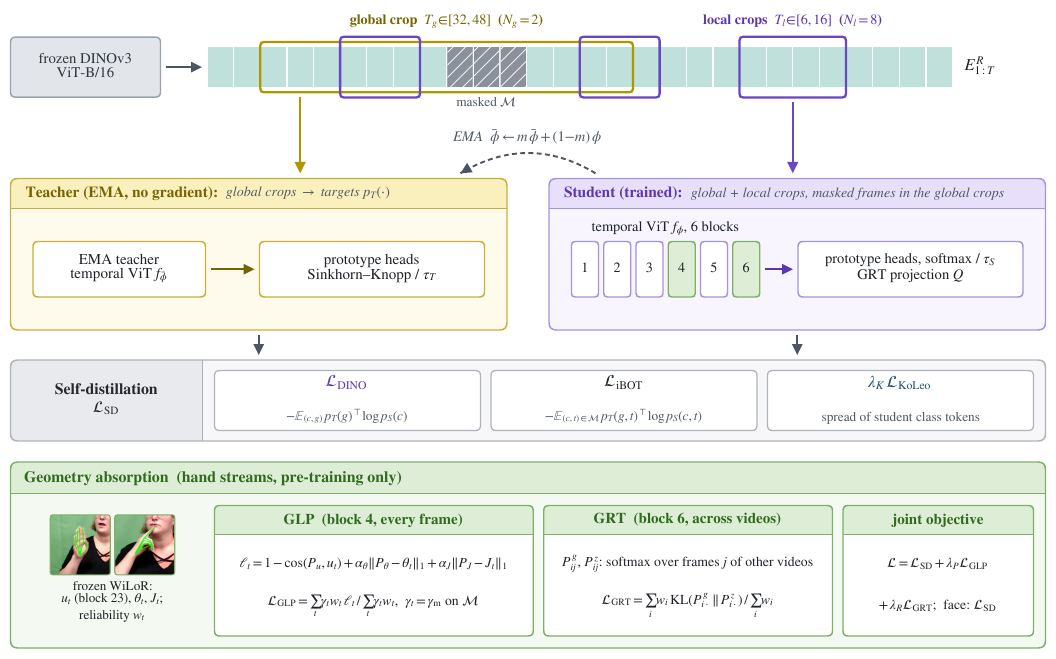}
\caption{Training of one hand stream. Long (global) and short (local) temporal crops are drawn from the cached \dinovthree\ features; the EMA teacher sees the global crops, and the student sees all crops with some frames of its global crops masked. Self-distillation combines the DINO, iBOT, and KoLeo terms. GLP reads student frame tokens at block~4 and GRT reads them at block~6, using targets from frozen WiLoR with reliability weighting applied to each loss; both losses are added from the first step. The face stream uses self-distillation alone.}
\label{fig:student_teacher}
\end{figure*}
\vspace{0.4em}

The output of a stream encoder should describe how the articulator evolves. We train it by temporal-axis self-distillation, which applies DINO's agreement between global and local views to the time axis (Figure~\ref{fig:student_teacher}). The student $f_\phi$ projects the sequence $(e_t)$ to width $D$, prepends a class token, adds sinusoidal encodings of the frame index, and applies $L$ pre-norm Transformer blocks; padded and undetected frames are excluded from attention. We write $z^{(l)}_t$ for the token of frame $t$ after block $l$. The teacher $f_{\bar\phi}$ is the exponential moving average (EMA) of the student.

Each training sample draws two global and several local temporal crops, whose lengths are sampled from a long and a short range. The teacher sees the global crops. The student sees all crops, and in its global crops a random subset $\mathcal{M}$ of frames has its input replaced by a learned mask embedding. A projection head maps class tokens and frame tokens to $K$ prototype scores. Writing $p_T(g)$ for the teacher's prototype distribution of global crop $g$, normalised over the batch by Sinkhorn--Knopp iterations~\citep{caron2020swav}, and $p_S(c)$ for the student's softmax distribution of crop $c$, the self-distillation objective combines a crop-level cross-entropy, its frame-level counterpart at masked frames (iBOT;~\citealp{zhou2021ibot}), and the KoLeo entropy regulariser on student class tokens~\citep{sablayrolles2018spreading}:
\begin{align}
\loss_{\mathrm{DINO}} &= -\mathbb{E}_{(c,g)}\big[p_T(g)^{\!\top}\!\log p_S(c)\big], \label{eq:l_dino}\\
\loss_{\mathrm{iBOT}} &= -\mathbb{E}_{(c,t)\in\mathcal{M}}\big[p_T(g,t)^{\!\top}\!\log p_S(c,t)\big], \label{eq:l_ibot}\\
\loss_{\mathrm{SD}} &= \loss_{\mathrm{DINO}} + \loss_{\mathrm{iBOT}} + \lambda_K \loss_{\mathrm{KoLeo}}, \label{eq:sd}
\end{align}
where the first expectation runs over pairs of a student crop and a different teacher global crop, and the second over masked frames of the student global crops paired with the same frames in the teacher.

\subsection{Geometric targets from a hand foundation model}
\label{sec:method:targets}

A frozen WiLoR~\citep{potamias2025wilor} supplies three targets for each available hand crop; left hands are mirrored into its right-hand convention. The geometric latent $u_t\in\mathbb{R}^{1280}$ is the mean of its 16 MANO pose-query tokens after backbone block $k$, standardised channel-wise using source-training statistics. The articulation $\theta_t\in\mathbb{R}^{15\times6}$ represents the 15 local finger-joint rotations of MANO~\citep{romero2017mano} in 6D~\citep{zhou2019rot6d}, excluding global wrist rotation. Its coordinate definition is independent of camera viewpoint. Wrist-relative joints $J_t\in\mathbb{R}^{21\times3}$ remain in the camera frame and retain orientation. These are WiLoR estimates, not motion-capture ground truth. Extraction and scale conventions are specified in Appendix~\ref{app:geometry_protocol}.

The latent should encode articulation while remaining predictable from appearance. We select $k{=}23$ by the product of two ridge-probe scores: articulation predicted from the latent, and the latent predicted from \dinovthree\ crop features. Table~\ref{tab:geometry_latents} reports the candidate-layer comparison: block 23 has the highest product, balancing articulatory information and learnability from appearance. Ridge fitting uses 10,000 source-training frames and evaluation uses 2,000 video-disjoint source-development frames, sampled equally from both hands. Appendix~\ref{app:geometry_protocol} gives the fitting details. Selection uses source-development videos; downstream test labels are not used.

\begin{table}[t]
\centering
\scriptsize
\setlength{\tabcolsep}{3pt}
\begin{tabular}{cccc}
\toprule
Block $k$ & $R^2(u\to\theta)$ & $R^2(e\to u)$ & Product \\
\midrule
7  & 0.54 & 0.72 & 0.389 \\
15 & 0.68 & 0.65 & 0.442 \\
23 & 0.79 & 0.59 & 0.466 \\
31 & 0.84 & 0.49 & 0.412 \\
\bottomrule
\end{tabular}
\caption{Latent-layer selection on source-development videos. Block 23 balances articulatory predictability and learnability from appearance.}
\label{tab:geometry_latents}
\end{table}

\paragraph{Reliability.}
Per-frame regressors fail on blurred and interacting hands, and such failures appear as abrupt jumps in $\theta$. We weight frame $t$ by
\begin{equation}
w_t = m_t\, o_t\, s_t\, \rho_t \exp\!\big(-\kappa\,\overline{|\theta_{t-1}-2\theta_t+\theta_{t+1}|}\big),
\label{eq:rel}
\end{equation}
where $o_t$ indicates that WiLoR produced valid estimates for the frame and both neighbours, $s_t$ is the detector score, and $\rho_t$ is 0.5 for a tracker-filled box and 1 otherwise. The bar averages coordinates of the second temporal difference of the 6D rotations. At video boundaries or missing-neighbour frames, $o_t{=}0$. Neighbours are adjacent frames of the downsampled source video, independently of temporal-crop boundaries. This is a temporal smoothness proxy for target reliability: both reconstruction jitter and genuine rapid changes can reduce it. Low-weight frames still contribute to self-distillation; a uniform-weight control is reported in \S\ref{sec:exp:ablations}.

\subsection{Geometric latent prediction}
\label{sec:method:glp}

GLP trains the student's intermediate block $l_g$ to encode the articulatory state of each frame. Lightweight heads $P_u$, $P_\theta$, $P_J$ ($384\to256\to d$ MLPs with GELU, $d\in\{1280,90,63\}$) read the frame tokens of the student's global crops after this block, giving a per-frame loss $\ell_t$ that is averaged with weights:
\begin{equation}
\begin{aligned}
\ell_t ={}& 1-\cos\big(P_u(z^{(l_g)}_t), u_t\big)\\
&+ \alpha_\theta\,\|P_\theta(z^{(l_g)}_t)-\theta_t\|_1\\
&+ \alpha_J\,\|P_J(z^{(l_g)}_t)-J_t\|_1,
\end{aligned}
\end{equation}
\begin{equation}
\loss_{\mathrm{GLP}} = \frac{\sum_t \gamma_t w_t\, \ell_t}{\max(\sum_t \gamma_t w_t,\epsilon)},
\label{eq:glp}
\end{equation}
where $\|\cdot\|_1$ averages coordinates, $\gamma_t{=}\gamma_{\mathrm{m}}{>}1$ for masked frames and 1 otherwise, and $\epsilon>0$ prevents an empty valid-target set from producing a division by zero. Empty sets contribute zero geometry loss. At a masked frame, the student infers the target from the visible temporal context. We supervise block $l_g{=}4$ of $L{=}6$, leaving two blocks above it for temporal self-distillation and GRT; Table~\ref{tab:abl_geometry} compares this placement with final-block supervision.

\subsection{Geometric relation transfer}
\label{sec:method:grt}

GLP supplies a target per frame. GRT additionally transfers relations between different videos at the last block. From the batch's global crops, we deduplicate frames by (video, stream, frame index), then uniformly sample up to $N{=}1024$ valid frames with $w_i>0$ across the distributed batch. Student descriptors are $\hat q_i=Q(z_i^{(L)})/\|Q(z_i^{(L)})\|$; geometric descriptors $\hat g_i$ concatenate separately $\ell_2$-normalised $u_i$ and $\theta_i$ and are normalised again. Let $\mathcal{N}(i)$ contain sampled frames from videos other than $i$'s. Anchors with no such candidates are omitted. With
\begin{equation}
P^{g}_{ij} = \frac{\exp(\hat g_i^{\top}\hat g_j/\tau_g)}{\sum_{j'\in\mathcal{N}(i)}\exp(\hat g_i^{\top}\hat g_{j'}/\tau_g)}
\end{equation}
and $P^{z}_{ij}$ defined in the same way from $\hat q$ with temperature $\tau_z$,
for $j\in\mathcal{N}(i)$, GRT minimises
\begin{equation}
\loss_{\mathrm{GRT}} = \frac{\sum_i w_i\,\mathrm{KL}\big(P^{g}_{i\cdot}\,\|\,P^{z}_{i\cdot}\big)}{\max(\sum_i w_i,\epsilon)}.
\label{eq:grt}
\end{equation}
Candidate distributions are unweighted; $w_i$ scales the KL term of anchor $i$. Excluding the anchor video removes immediate temporal neighbours, making the objective transfer geometric relations beyond one trajectory. Video identity is available without signer labels; different videos can contain the same signer. GRT therefore imposes cross-video, rather than guaranteed cross-signer, supervision. Its relation descriptors emphasise articulation; the GLP joints retain camera-relative orientation. We evaluate cross-signer handshape grouping separately using human phonological labels (\S\ref{sec:exp:phonology}).

\subsection{Objective and inference}
\label{sec:method:obj}

The two hand streams are trained with
\begin{equation}
\loss = \loss_{\mathrm{SD}} + \lambda_P\,\loss_{\mathrm{GLP}} + \lambda_R\,\loss_{\mathrm{GRT}},
\label{eq:total}
\end{equation}
and the face stream with $\loss_{\mathrm{SD}}$ alone. All geometry targets are fixed and all terms are active from the first update. The inference encoders are the three EMA teachers; WiLoR and $P_u,P_\theta,P_J,Q$ are discarded. Thus geometry adds no inference modules to the same temporal appearance backbone. A task-specific learned softmax mixes the teacher's $L$ block outputs and its projected \dinovthree\ input~\citep{yang2021superb}; the three stream features are projected and concatenated for the downstream head. Offline feature caching accelerates training, while processing a new video still requires detection, tracking, and the image encoder.

\section{Experiments}
\label{sec:experiments}

We follow \shubert's public-data evaluation suite for source-corpus accounting, target splits, task heads, metrics, and downstream schedules. Published baselines provide task-level context; the within-backbone controls in Table~\ref{tab:abl_geometry} isolate geometric supervision. The CVPR \signdinobaseline~\citep{gan2026signdino} is a separate frame-level method. Appendix~\ref{app:repro} gives the training protocol; Appendix~\ref{app:geometry_protocol} gives geometry diagnostics.

\subsection{Pre-Training Setup}
\label{sec:exp:setup}

We pre-train \ours\ on the YouTube-ASL source split of \shubert~\citep{uthus2023youtubeasl,gueuwou2025shubert}, in which clips intersecting the OpenASL evaluation set are replaced with non-overlapping ASL videos from YouTube-SL-25~\citep{tanzer_zhang2024ytsl25} (${\approx}984$ source hours). Each stream has a 10.8M-parameter temporal Transformer ($L{=}6$, $D{=}384$, 6 heads) on the shared frozen \dinovthree\ ViT-B/16. Temporal encoders and DINO/iBOT heads comprise approximately 47M trainable parameters; auxiliary geometry heads are accounted separately in Appendix~\ref{app:geometry_protocol}. All runs use $4\times$A100 80\,GB GPUs; training schedules and compute are in Appendices~\ref{app:repro} and~\ref{app:compute}.

\subsection{Sign Language Translation}
\label{sec:exp:slt}

The translation head replicates \shubert\ \S4.2: the layer-weighted teacher features of each stream are normalised, projected, concatenated, and fed to a ByT5-Base decoder~\citep{xue2022byt5}. Training has two stages (Table~\ref{tab:slt_train}): a source stage on the YouTube-ASL split above and a target stage on How2Sign or OpenASL; FLEURS-ASL is scored zero-shot after the source stage. The \emph{live} variant updates the temporal encoders at $1/10$ of the decoder learning rate, and the \emph{frozen} variant trains only the fusion projection and ByT5. We report BLEU-4~\citep{papineni2002bleu} via SacreBLEU~\citep{post2018sacrebleu} and BLEURT-20~\citep{sellam2020bleurt}.

\begin{table*}[t]
\centering
\scriptsize
\setlength{\tabcolsep}{3.2pt}
\resizebox{\textwidth}{!}{%
\begin{tabular}{lccccccc}
\toprule
\multirow{2}{*}{Method} & \multirow{2}{*}{SSL} & \multirow{2}{*}{PT hrs} & \multicolumn{2}{c}{How2Sign} & \multicolumn{2}{c}{OpenASL} & FLEURS-ASL \\
\cmidrule(lr){4-5}\cmidrule(lr){6-7}\cmidrule(lr){8-8}
& & & BLEU$\uparrow$ & BLEURT$\uparrow$ & BLEU$\uparrow$ & BLEURT$\uparrow$ & BLEU$\uparrow$ \\
\midrule
SSVP-SLT~\citep{rust2024ssvpslt} & \checkmark & 1054 & 15.5 & 49.6 & --- & --- & --- \\
YouTube-ASL fine-tune~\citep{uthus2023youtubeasl} & $\times$ & 984 & 12.4 & 46.6 & --- & --- & --- \\
YouTube-SL-25~\citep{tanzer_zhang2024ytsl25} & $\times$ & 3207 & 15.4 & 47.9 & --- & --- & 4.4 \\
SignMusketeers~\citep{gueuwou2025signmusketeers} & \checkmark & 984 & 14.3 & --- & --- & --- & --- \\
SpaMo~\citep{hwang2025spamo} & $\times$ & --- & 10.11 & 42.23 & --- & --- & --- \\
Uni-Sign~\citep{li2025unisign} & $\times$ & 984 & 14.9 & 49.4 & 23.1 & 60.4 & --- \\
OpenASL~\citep{shi2022openasl} & $\times$ & --- & --- & --- & 6.7 & 31.1 & --- \\
\shubert~\citep{gueuwou2025shubert} & \checkmark & 984 & 16.2 & 49.9 & 23.2 & 60.6 & 4.7 \\
\signdinobaseline~\citep{gan2026signdino} & \checkmark & --- & 15.47 & --- & 17.46 & --- & --- \\
\midrule
\multicolumn{8}{c}{\textit{\ours}} \\
\ours\ frozen + ByT5 (frozen encoders) & \checkmark & 984 & $16.8_{\pm 0.3}$ & $50.4_{\pm 0.4}$ & $23.6_{\pm 0.3}$ & $60.9_{\pm 0.3}$ & $4.9_{\pm 0.2}$ \\
\ours\ live fine-tune (updated encoders) & \checkmark & 984 & $\mathbf{17.9_{\pm 0.15}}$ & $\mathbf{51.3_{\pm 0.4}}$ & $\mathbf{24.5_{\pm 0.3}}$ & $\mathbf{61.7_{\pm 0.3}}$ & $\mathbf{5.3_{\pm 0.2}}$ \\
\ours\ live, last-layer only (no weighted sum) & \checkmark & 984 & $15.8_{\pm 0.3}$ & $48.7_{\pm 0.4}$ & $23.0_{\pm 0.3}$ & $59.7_{\pm 0.4}$ & $4.6_{\pm 0.2}$ \\
\bottomrule
\end{tabular}%
}
\caption{Sign language translation under the \shubert\ public-data protocol. PT hrs: source pre-training video hours. \ours\ is pre-trained on the OpenASL-decontaminated YouTube-ASL split, fine-tuned on How2Sign or OpenASL, and evaluated zero-shot on FLEURS-ASL; Frozen/live denote whether the temporal encoders are updated; this is the freezing convention of \shubert, not equivalence to its last-layer ablation numbers. \ours\ entries are mean $\pm$ std over 3 seeds; baselines are published results. Dashes indicate values not reported in the cited work.}
\label{tab:slt}
\vspace{-0.4em}
\end{table*}

\paragraph{Results.}
Live-fine-tuned \ours\ reaches \emph{17.9}/\emph{24.5} BLEU-4 on How2Sign/OpenASL (Table~\ref{tab:slt}), \emph{+1.7}/\emph{+1.3} BLEU-4 and \emph{+1.4}/\emph{+1.1} BLEURT over \shubert, and \emph{+2.43}/\emph{+7.04} BLEU-4 over the published \signdinobaseline\ results. With frozen encoders it reaches 16.8/23.6, exceeding the published live \shubert\ result. WiLoR and the geometry heads are absent at test time. \ours\ also exceeds Uni-Sign on OpenASL (24.5 vs.\ 23.1) and \shubert\ on zero-shot FLEURS-ASL (5.3 vs.\ 4.7). Reading only the last temporal layer instead of the layer-weighted sum lowers How2Sign to 15.8 (14.7 frozen; Table~\ref{tab:abl_layer}), indicating that multiple encoder depths contribute to decoding.

\subsection{Isolated Sign Language Recognition}
\label{sec:exp:islr}

We evaluate ISLR on ASL Citizen~\citep{desai2023aslcitizen}, Sem-Lex~\citep{kezar2023semlex}, and WLASL2000~\citep{li2020wlasl}, the \shubert\ suite (MSASL is excluded because its test set overlaps YouTube-ASL). The head follows \shubert\ \S4.3: a stream-fused, mask-aware temporal average followed by BatchNorm and a linear classifier. We report three regimes: frozen features with a linear head, rank-1 LoRA~\citep{hu2022lora} on the temporal encoders (0.11M parameters), and full fine-tuning of the temporal encoders.

\begin{table*}[t]
\centering
\scriptsize
\setlength{\tabcolsep}{3pt}
\resizebox{\textwidth}{!}{%
\begin{tabular}{lcccccccccc}
\toprule
\multirow{2}{*}{Method} & \multirow{2}{*}{Trainable} & \multicolumn{3}{c}{ASL\,Citizen} & \multicolumn{3}{c}{Sem-Lex} & \multicolumn{2}{c}{WLASL2000} & ASL-STEM-Wiki \\
\cmidrule(lr){3-5}\cmidrule(lr){6-8}\cmidrule(lr){9-10}\cmidrule(lr){11-11}
& & R@1$\uparrow$ & R@5$\uparrow$ & R@10$\uparrow$ & R@1$\uparrow$ & R@5$\uparrow$ & R@10$\uparrow$ & P-I$\uparrow$ & P-C$\uparrow$ & mIoU$\uparrow$ \\
\midrule
\multicolumn{11}{c}{\textit{Baselines}} \\
ST-GCN~\citep{desai2023aslcitizen} & 0.45M & 0.60 & 0.82 & 0.88 & --- & --- & --- & --- & --- & --- \\
SignCLIP~\citep{jiang2024signclip} & 217M & 0.60 & 0.84 & 0.89 & 0.30 & 0.48 & 0.55 & --- & --- & --- \\
I3D~\citep{desai2023aslcitizen} & 25M & 0.63 & 0.86 & 0.91 & --- & --- & --- & --- & --- & --- \\
SignBERT+~\citep{hu2023signbertplus} & --- & --- & --- & --- & --- & --- & --- & 48.85 & 46.37 & --- \\
MSLU~\citep{zhou2024mslu} & --- & --- & --- & --- & --- & --- & --- & 56.29 & 53.29 & --- \\
Uni-Sign~\citep{li2025unisign} & 580M & --- & --- & --- & --- & --- & --- & 63.52 & 61.32 & --- \\
\shubert\ (rank-1 LoRA)~\citep{gueuwou2025shubert} & 0.17M & 0.65 & 0.87 & 0.91 & 0.54 & 0.74 & 0.80 & 60.90 & 58.01 & 0.40 \\
\midrule
\multicolumn{11}{c}{\textit{\ours}} \\
\ours\ frozen + linear & linear & $0.63_{\pm .01}$ & $0.86_{\pm .01}$ & $0.91_{\pm .01}$ & $0.52_{\pm .01}$ & $0.72_{\pm .01}$ & $0.79_{\pm .01}$ & $60.1_{\pm 0.4}$ & $57.5_{\pm 0.4}$ & $0.39_{\pm .01}$ \\
\ours\ rank-1 LoRA & 0.11M & $0.67_{\pm .01}$ & $0.88_{\pm .01}$ & $0.92_{\pm .01}$ & $0.56_{\pm .01}$ & $0.76_{\pm .01}$ & $0.82_{\pm .01}$ & $66.9_{\pm 0.4}$ & $64.2_{\pm 0.4}$ & $0.41_{\pm .01}$ \\
\ours\ full fine-tune & all & $\mathbf{0.704_{\pm .01}}$ & $\mathbf{0.89_{\pm .01}}$ & $\mathbf{0.93_{\pm .01}}$ & $\mathbf{0.593_{\pm .01}}$ & $\mathbf{0.78_{\pm .01}}$ & $\mathbf{0.84_{\pm .01}}$ & $\mathbf{68.5_{\pm 0.4}}$ & $\mathbf{65.9_{\pm 0.4}}$ & $\mathbf{0.43_{\pm .01}}$ \\
\bottomrule
\end{tabular}%
}
\caption{Isolated sign recognition (Recall@$k$; WLASL2000 per-instance/per-class top-1) and fingerspelling detection (ASL-STEM-Wiki mIoU) under the \shubert\ suite; Sem-Lex uses \shubert's full test set. \ours: mean $\pm$ std over 3 seeds. Dashes indicate values not reported in the cited work.}
\label{tab:islrfs}
\vspace{-0.4em}
\end{table*}

\paragraph{Fingerspelling detection.}
On ASL-STEM-Wiki~\citep{yin2024aslstemwiki}, detection is per-frame binary classification with a two-layer Transformer over the fused features, followed by thresholding, median smoothing, and interval extraction; we report mean interval-IoU under the dataset's cross-validation protocol (Appendix~\ref{app:repro}).

\paragraph{Results.}
With rank-1 LoRA (Table~\ref{tab:islrfs}), \ours\ improves Recall@1 over \shubert\ by \emph{+0.02} on both ASL Citizen (\emph{0.67}) and Sem-Lex (\emph{0.56}); full fine-tuning reaches \emph{0.704} and \emph{0.593}. On WLASL2000, LoRA gives \emph{66.9}/\emph{64.2} P-I/P-C, above Uni-Sign (\emph{63.52}/\emph{61.32}) and \shubert\ (\emph{60.90}/\emph{58.01}), and full fine-tuning gives \emph{68.5}/\emph{65.9}. Fingerspelling mIoU is \emph{0.41} with LoRA (\shubert\ \emph{0.40}) and \emph{0.43} with full fine-tuning. Frozen features with a linear head reach \emph{0.63}/\emph{0.52} R@1 on ASL Citizen/Sem-Lex; the gains over \shubert\ are obtained with matched rank-1 adaptation.

\subsection{Phonological and Articulatory Probing}
\label{sec:exp:phonology}

We train 16 classifiers for ASL-Lex 2.0 categories following \shubert\ (Appendix~\ref{app:phonology}). Mean R@1 is 87.64/91.85 on Sem-Lex/ASL Citizen, compared with 79.38/85.02 for \shubert. On Sem-Lex, Handshape improves from 62.93 to 71.16, Selected Fingers from 79.53 to 89.94, and Spread from 79.42 to 90.99. These labels are independent of WiLoR. A frozen ridge probe on How2Sign explains 56\% of the variance in WiLoR-estimated articulation versus 44\% for raw \dinovthree\ features, measuring transfer of the teacher's geometric estimates. A frozen cross-signer retrieval diagnostic additionally tests human-labelled handshape across different lexical signs; its results are in Table~\ref{tab:geometry_retrieval}.

\paragraph{Frozen cross-signer handshape retrieval.}
We use ASL Citizen's official signer-disjoint test queries and training-gallery clips that map to a valid ASL-Lex 2.0 dominant-handshape label. No model is fine-tuned for this diagnostic. We average valid hand-stream frame features (block 4 for temporal models), concatenate the two hand streams, and $\ell_2$-normalise. We rank gallery clips by cosine similarity, excluding the query signer and query lexical sign. P@5 is the fraction of the five retrieved clips with the same human-assigned handshape, averaged first within handshape and then across supported handshapes. We use all eligible queries and at most ten gallery clips per lexical sign, sampled deterministically with seed 0. This tests shared articulation across signers and lexical items; it does not test calibrated multi-view geometry.

Table~\ref{tab:geometry_retrieval} reports P@5 of 0.61 for full \ours, compared with 0.41 for raw \dinovthree\ and 0.49 for SD only. GLP alone reaches 0.55, while adding GRT further improves grouping across signers and lexical signs.

\begin{table}[t]
\centering
\scriptsize
\setlength{\tabcolsep}{3pt}
\begin{tabular}{lcc}
\toprule
Frozen representation & Handshape P@5 & Articulation $R^2$ \\
\midrule
Raw \dinovthree & 0.41 & 0.44 \\
SD only & 0.49 & 0.48 \\
SD + GLP (w/o GRT) & 0.55 & 0.54 \\
Full \ours & 0.61 & 0.56 \\
\bottomrule
\end{tabular}
\caption{Frozen geometry diagnostics. Handshape retrieval uses independent human labels and excluded-signer/excluded-lexeme candidates; articulation $R^2$ predicts WiLoR estimates on How2Sign.}
\label{tab:geometry_retrieval}
\end{table}

\subsection{Ablations and Representation Diagnostics}
\label{sec:exp:ablations}

Table~\ref{tab:abl_geometry} isolates GLP and GRT within the same architecture and training schedule. Table~\ref{tab:abl_ours} retains temporal-context and tracking controls; Table~\ref{tab:abl_streams} removes articulator streams. Masking, data scale, layer pooling, backbone, decoder, and crop-length comparisons appear in Appendix~\ref{app:moreabl}, and substitutions of the hand detector and the hand geometry model in Appendix~\ref{app:handmodel}.

\begin{table}[t]
\centering
\scriptsize
\setlength{\tabcolsep}{3pt}
\begin{tabularx}{\columnwidth}{p{3.3cm}C}
\toprule
Streams kept at fusion head & BLEU \\
\midrule
LH + Face   (no RH)               & 9.8  \\
RH + Face   (no LH)               & 11.2 \\
LH + RH     (no Face)             & 14.6 \\
LH + RH + Face (default)          & 17.9 \\
\bottomrule
\end{tabularx}
\caption{Leave-one-stream-out at the fusion head, How2Sign test BLEU-4 (live fine-tuning; SSL unchanged).}
\label{tab:abl_streams}
\vspace{-0.6em}
\end{table}
\vspace{0.4em}

\paragraph{Stream contribution.}
Dropping the right-hand stream costs the most (\emph{-8.1}~BLEU from 17.9), followed by the left hand (\emph{-6.7}) and the face (\emph{-3.3}), in line with the predominantly right-handed signers of YouTube-ASL and How2Sign; the face adds mouthing and non-manual cues.

\paragraph{Geometry at fixed backbone and schedule.}
Removing GRT while retaining GLP lowers How2Sign from 17.9 to 16.5 BLEU and ASL Citizen from 0.704 to 0.670 R@1. Relative to SD only (15.6 BLEU, 0.642 R@1), GLP and GRT separately improve both tasks, and their combination reaches the highest scores. Moving GLP to block 6 gives 17.2 BLEU and 0.690 R@1, supporting intermediate supervision with temporal layers above it. All controls hold the image backbone, crops, encoder capacity, and downstream heads fixed.

\begin{table}[t]
\centering
\scriptsize
\setlength{\tabcolsep}{3pt}
\begin{tabularx}{\columnwidth}{Ycc}
\toprule
Objective / placement & H2S BLEU & ASLC R@1 \\
\midrule
SD only & 15.6 & 0.642 \\
SD + GLP (w/o GRT) & 16.5 & 0.670 \\
SD + GRT (w/o GLP) & 16.4 & 0.665 \\
SD + GLP + GRT (block 4) & 17.9 & 0.704 \\
SD + GLP + GRT (block 6) & 17.2 & 0.690 \\
\bottomrule
\end{tabularx}
\caption{Matched geometry controls: identical 984-hour source split, 450K updates, and downstream schedules. H2S uses live SLT; ASLC uses full temporal fine-tuning.}
\label{tab:abl_geometry}
\vspace{-0.4em}
\end{table}

\paragraph{Reliability-weighted articulation probes.}
Using the same valid How2Sign fitting frames, reliability-weighted ridge regression with Eq.~\eqref{eq:rel} reaches $R^2{=}0.56$, compared with 0.53 under uniform weights. This offline control evaluates how target weighting affects probe fitting, with the SSL encoders fixed. Appendix~\ref{app:geometry_protocol} gives the fitting details.

\paragraph{Temporal context and tracking.}
Restricting the teacher to short windows lowers H2S BLEU/ASLC R@1 to 15.8/0.65; removing ByteTrack gives 16.4/0.61 (Table~\ref{tab:abl_ours}). The tracked streams and long temporal context contribute alongside geometry.

\begin{table}[t]
\centering
\scriptsize
\setlength{\tabcolsep}{3pt}
\begin{tabularx}{\columnwidth}{p{3.4cm}CC}
\toprule
Ablation & H2S BLEU & ASLC R@1 \\
\midrule
Full \ours & 17.9 & 0.704 \\
w/o GRT ($\lambda_R{=}0$) & 16.5 & 0.67 \\
Teacher short window only & 15.8 & 0.65 \\
YOLO only (no ByteTrack) & 16.4 & 0.61 \\
\bottomrule
\end{tabularx}
\caption{Ablations of GRT ($\lambda_R{=}0$ in Eq.~\eqref{eq:total}, GLP kept), teacher temporal context, and tracking.}
\label{tab:abl_ours}
\vspace{-0.6em}
\end{table}

\paragraph{Additional design-axis and protocol ablations.}
The frame-only mask gives the best BLEURT (Table~\ref{tab:abl_mask}); the randomised crop-length sampler beats fixed windows by \emph{0.4}--\emph{0.6}~BLEU (Table~\ref{tab:abl_windowsize}); How2Sign BLEU spans \emph{16.5}--\emph{18.0} over five frozen DINO backbones (Table~\ref{tab:abl_backbone}); and \ours\ beats \shubert\ by \emph{+1.4}--\emph{+1.7}~BLEU across five decoders (Table~\ref{tab:abl_decoder}). Going from 10\% to 100\% of the pre-training data adds \emph{+3.3}~BLEU.

\paragraph{Training and downstream diagnostics.}
Prototype-use entropy remains above 0.91; task-specific layer weights spread across more depths for SLT than for recognition and fingerspelling. Appendix~\ref{app:diagnostics} interprets these as prototype-occupancy and layer-use diagnostics, and Appendix~\ref{app:clusters} provides descriptive clusters.

\section{Conclusion}
We presented \sdeg, which transfers hand-model targets into an intermediate temporal layer and geometric frame relations into the final layer. The geometry model is used only for pre-training. Across ASL translation, recognition, fingerspelling, and phonological classification, the resulting representation improves over the published \shubert\ task results in matched adaptation regimes. The GRT removal control costs 1.4 BLEU-4 on How2Sign and 0.034 R@1 on ASL Citizen; matched geometry controls and phonological retrieval assess the additional roles of GLP, layer placement, and cross-video relations.

\section*{Limitations}
\ours\ depends on hand and face crops: prolonged detector or tracker dropouts remove visual evidence. WiLoR provides estimated geometry and imports the biases and errors of its supervised training; local articulation coordinates do not guarantee view-invariant estimates. The temporal smoothness weight may also downweight rapid signing. GRT excludes same-video neighbours but does not require signer identity during pre-training. The human-labelled retrieval diagnostic tests cross-signer handshape grouping, while calibrated cross-view reconstruction is outside this evaluation. Experiments cover ASL and ASL-to-English, and the three-stream model does not explicitly encode upper-body pose. Privacy, consent, and Deaf-community participation remain important when extending to other languages or deploying these representations~\citep{desai2024_systemic_biases}. New-video inference still includes the detector, tracker, and image backbone, and hand-target extraction adds pre-training cost.

\bibliography{references}

\appendix
\section{\sdeg\ Pre-training Hyperparameters}
\label{app:sdeg_hparams}

\begin{table}[h]
\centering
\small
\resizebox{\columnwidth}{!}{%
\begin{tabular}{ll}
\toprule
Parameter & Value \\
\midrule
Frame rate & 12.5 fps \\
Crop size / padding & $224^2$; hands $1.4\times$, face $1.2\times$ \\
Temporal encoder & $L{=}6$, $D{=}384$, 6 heads \\
Global / local crops & $2\times[32,48]$ / $8\times[6,16]$ frames \\
Masked fraction & $\mathcal{U}[0.25,0.50]$ \\
Prototypes $K$ & 8192 \\
Teacher / student temperature & 0.04$\to$0.07, then 0.07 / 0.1 \\
$\lambda_K$ & 0.1 \\
$\lambda_P$, $\lambda_R$ & 1.0, 1.0 \\
$\alpha_\theta$, $\alpha_J$, $\gamma_{\mathrm{m}}$ & 1.0, 1.0, 2.0 \\
$l_g$, $k$ (probed: 7, 15, 23, 31) & 4, 23 \\
GRT: $N$, $\tau_g$, $\tau_z$ & 1024, 0.05, 0.1 \\
Reliability $\kappa$, $\rho$ & 3.0, 0.5 \\
Downstream projection per stream & 256 \\
Optimiser & AdamW, lr $5{\times}10^{-4}$ then $2{\times}10^{-4}$, wd $0.04\to0.4$ \\
EMA momentum & $0.994\to1.0$, then $0.999\to1.0$ \\
Batch / updates & 32 videos / 400K + 50K (Table~\ref{tab:pretrain}) \\
\bottomrule
\end{tabular}}
\caption{\sdeg\ pre-training hyperparameters (all streams; geometry terms on the hand streams only).}
\end{table}

\section{Geometric Target and Diagnostic Protocols}
\label{app:geometry_protocol}

\paragraph{Extraction conventions.}
WiLoR's 32-block, 1280-dimensional ViT includes 16 MANO pose-query tokens. Local rotations exclude the wrist's global rotation; wrist-relative joints retain the camera axes. We use the released WiLoR checkpoint and native crop normalisation, read backbone blocks with zero-based indices 7/15/23/31, and extract the refined MANO articulation and joints. We mirror left-hand input horizontally into the right-hand convention, keep local finger rotations in that canonical convention, and reflect wrist-relative joints back to original camera axes with $S=\operatorname{diag}(-1,1,1)$. We express $J$ in metres and fit per-channel latent statistics on source-training frames only ($\epsilon=10^{-6}$).

\paragraph{Implementation specification.}
We use the existing custom YOLOv8n hand/face checkpoint. Its weight hash, category mapping, and detector-training manifest will be released with the configurations. For a frontal signer, we initialise anatomical left/right tracks from image-right/image-left respectively and retain their ByteTrack IDs through crossing motions. Cases with ambiguous initial side assignment are flagged in the validity mask. Tracker-filled boxes retain the last detector score; geometry neighbours are adjacent frames of the original downsampled video, not temporal-crop boundaries. For GRT, we deduplicate overlapping global-crop frames by (video, stream, frame index), then uniformly sample up to 1024 valid frames across the distributed batch; we remove same-video candidates and omit anchors with no remaining candidates. Candidate distributions are unweighted; Eq.~\eqref{eq:grt} weights anchors.

Each GLP head is $384\to256\to d$, with GELU and $d\in\{1280,90,63\}$; $Q$ is a linear $384\to128$ projection. Separate DINO/iBOT heads use $384\to512\to256\to8192$, with GELU in hidden layers and a normalised bottleneck before prototypes. Geometry heads add approximately 1.4M student parameters across the two hand streams (approximately 48.4M total trained parameters with the 47M temporal/prototype core); they are not EMA inference modules. Downstream inference uses full attention over each benchmark clip, with padded/invalid frames masked and omitted from temporal averaging.

\paragraph{Latent selection and ridge probes.}
We sample 10,000 valid source-training frames and 2,000 source-development frames, with video-disjoint splits and equal sampling from the two hands. We fit ridge regressions $u^{(k)}\!\to\theta$ and $e\!\to u^{(k)}$ after standardising predictors on the fitting split. We select the regularisation coefficient from $\{10^{-4},10^{-3},10^{-2},10^{-1},1,10\}$ by a video-disjoint inner split. We average coordinate-wise test $R^2$ and multiply the two development scores to select $k$. Table~\ref{tab:geometry_latents} reports the scores; no downstream test videos enter layer selection. The How2Sign articulation probes yield 44\%/56\% explained variance, with the same fitting procedure applied on How2Sign train/validation/test respectively.

The frozen cross-signer handshape-retrieval protocol and results are reported in \S\ref{sec:exp:phonology} and Table~\ref{tab:geometry_retrieval}.

\paragraph{Reliability and cost.}
We stratify retained target weight by motion-speed quartile to check suppression of rapid valid signing. WiLoR target extraction adds approximately 160 A100 GPU-hours for both hands; combined pre-training plus appearance caching and geometry extraction is then approximately 1300 GPU-hours. These costs are reported in GPU-hours; the \shubert\ figure uses different hardware.

\section{Default Hyperparameter Card}
\label{app:hyperparams}

Table~\ref{tab:hyperparams} is the high-level configuration card; the SSL optimisation segments (Table~\ref{tab:pretrain}) and the source/target translation training schedule (Table~\ref{tab:slt_train}) are reported in Appendix~\ref{app:repro}.

\begin{table}[h]
\centering
\scriptsize
\setlength{\tabcolsep}{3pt}
\begin{tabularx}{\columnwidth}{p{2.4cm}Y}
\toprule
Hyperparameter & Default value \\
\midrule
Backbone & frozen \dinovthree\ ViT-B/16 \\
Streams & LH, RH, face (3 streams, independent SSL) \\
Temporal encoder & $L{=}6$, $D{=}384$, 6 heads, MLP ratio 4 \\
Crops (contiguous) & $T_g\!\sim\!\mathcal{U}\{32{-}48\}$, $T_l\!\sim\!\mathcal{U}\{6{-}16\}$; $N_g{=}2$, $N_l{=}8$ \\
Temperatures & $\tau_T:0.04\!\to\!0.07$, $\tau_S=0.1$; Table~\ref{tab:pretrain} \\
EMA momentum & cosine, $0.994\to1.0$ then $0.999\to1.0$ (Table~\ref{tab:pretrain}) \\
Prototype $K$ & $8192$ \\
Geometry weights $\lambda_P,\lambda_R$ & $1.0,\ 1.0$ \\
KoLeo weight & $0.1$ \\
Optimiser & AdamW, cosine decay, bf16 \\
Detector + tracker & YOLOv8n + ByteTrack; gaps $\leq3$ frames filled \\
Downstream SLT & ByT5-Base, beam $=5$, $L_{\max}{=}384$, label smoothing $0.2$ \\
Downstream ISLR & BN + linear; rank-1 LoRA; 125 epochs, batch 128 \\
Downstream FS & 2-layer Transformer + frame head; threshold $0.5$ \\
\bottomrule
\end{tabularx}
\caption{Default hyperparameter card. Values summarise the default configuration used for the main experiments and ablations.}
\label{tab:hyperparams}
\end{table}
\vspace{0.4em}

\paragraph{EMA momentum schedule under variable-length crops.}
EMA momentum follows the per-update cosine schedules in Table~\ref{tab:pretrain}. Each video supplies $N_gT_g+N_lT_l\in[112,224]$ student frame tokens, with expectation 168. DINO averages crop pairs, iBOT averages valid masked positions, and KoLeo averages class tokens. These normalisations avoid scaling the loss directly with crop length; they do not imply identical gradient variance across crop samplers. Geometry losses are normalised by their valid target weights.

\section{Qualitative Cluster Analysis of \ours\ Features}
\label{app:clusters}

\begin{figure}[!h]
\centering
\includegraphics[width=\columnwidth]{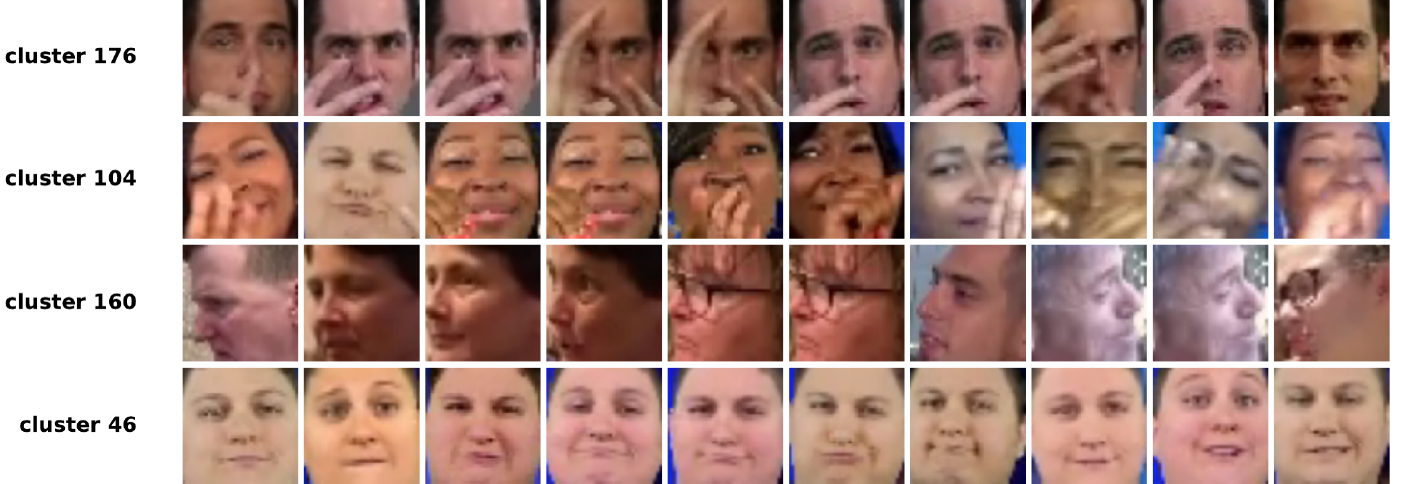}
\caption{Sample \emph{face} clusters from \ours\ features on held-out How2Sign and OpenASL frames. Four $k$-means clusters ($K{=}250$); each row shows 10 nearest-to-centroid members. Cluster~176 captures ``hand-on-face mouthings'' across multiple identities; cluster~104 captures hand-near-mouth gestures with varied skin tones; cluster~160 surfaces side-profile / looking-down poses; cluster~46 collects neutral frontal views with a closed mouth.}
\label{fig:clusters_face}
\end{figure}

\begin{figure}[!h]
\centering
\includegraphics[width=\columnwidth]{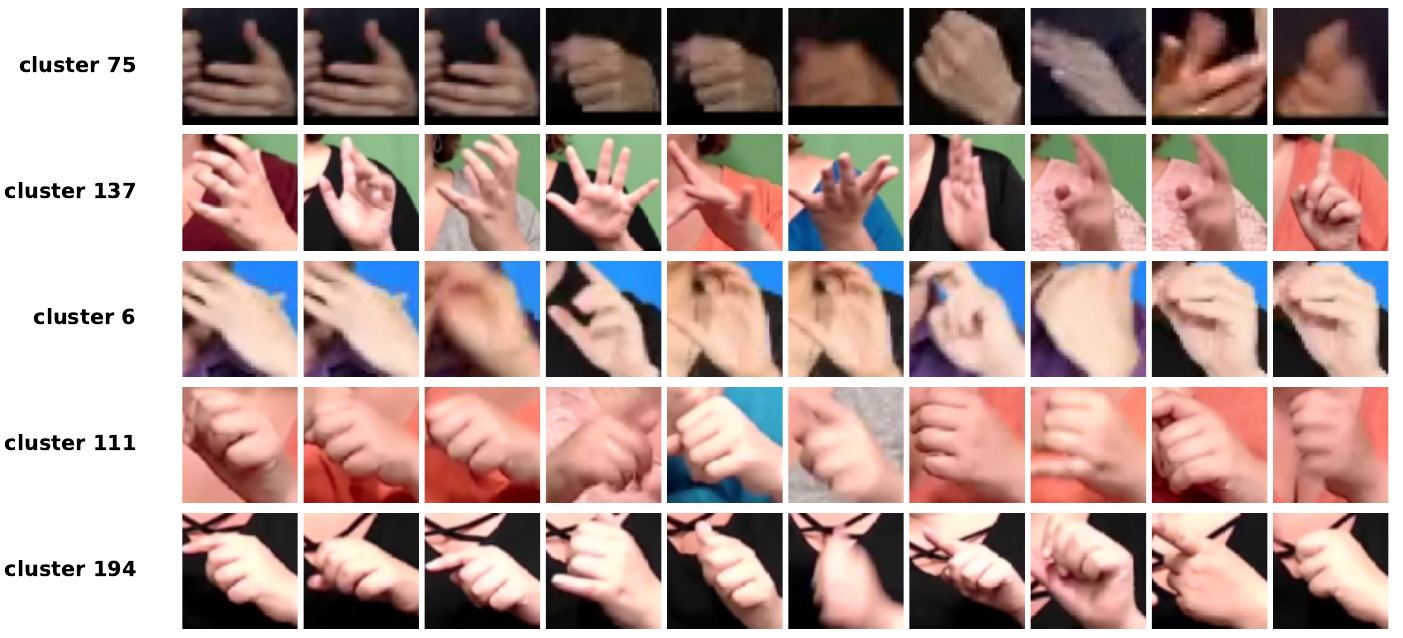}
\caption{Sample \emph{left-hand} clusters from \ours\ features on held-out How2Sign and OpenASL frames. Five $k$-means clusters ($K{=}200$). Cluster~75 captures thumb-up / closed-fist handshapes; cluster~137 captures the spread-fingers (``5'') handshape across multiple backgrounds; clusters~6 and 111 capture mid-motion signing gestures; cluster~194 captures two-hand contact poses.}
\label{fig:clusters_lh}
\end{figure}

\begin{figure}[!h]
\centering
\includegraphics[width=\columnwidth]{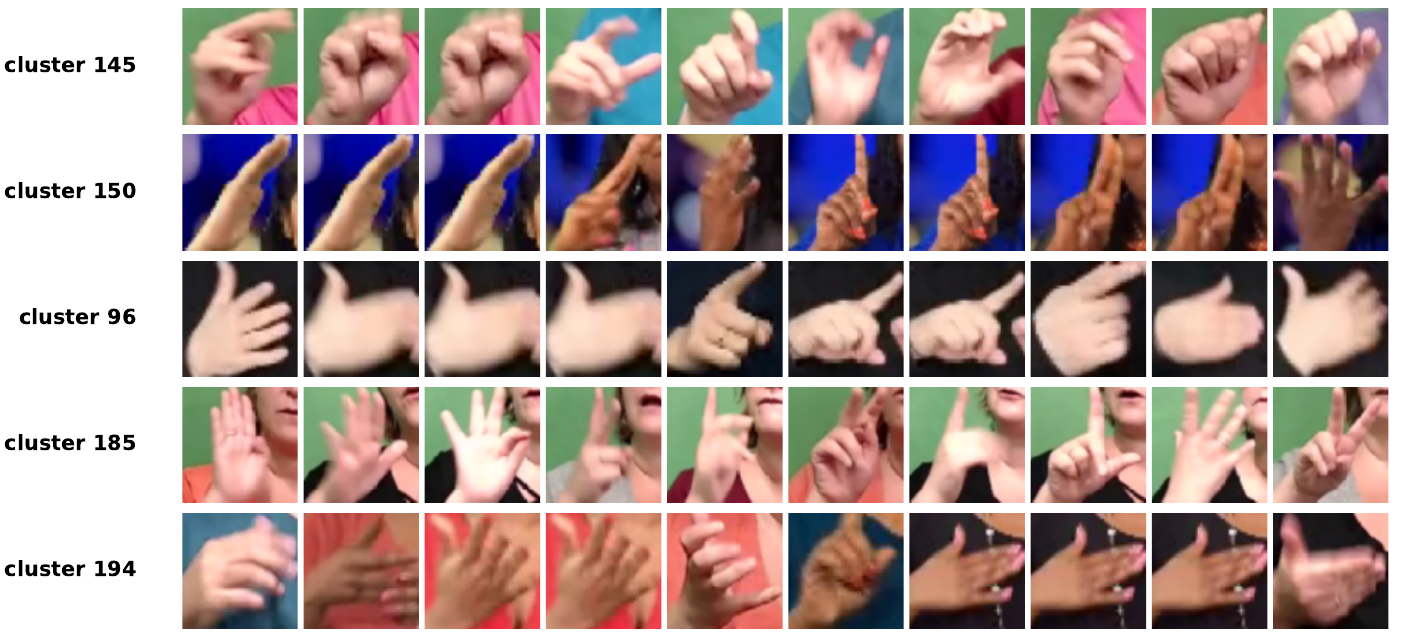}
\caption{Sample \emph{right-hand} clusters from \ours\ features on held-out How2Sign and OpenASL frames. Five $k$-means clusters ($K{=}200$). Cluster~145 captures pointing / extended-index handshapes across multiple backgrounds; cluster~150 captures a related forward-pointing pose on dark backgrounds; cluster~96 captures the thumb-up / closed-fist family; cluster~185 captures open-palm / spread-fingers gestures; cluster~194 captures motion-blurred active gesturing.}
\label{fig:clusters_rh}
\end{figure}

To examine what the \ours\ encoder captures, we run a fully zero-shot qualitative analysis on held-out frames, mirroring the cluster-visualisation protocol of \shubert\ Appendix~B (Figures~5--7 there)~\citep{gueuwou2025shubert}. After the SSL training described in \S\ref{sec:method:ssl} --- no signing gloss or translation labels, and with OpenASL clips that overlap with the YouTube-ASL pre-training source already decontaminated at the SSL stage (\S\ref{sec:exp:setup}) --- we apply the frozen \ours\ pipeline to a fresh sample of \emph{previously unseen} How2Sign and OpenASL videos: YOLOv8n + ByteTrack extracts synchronised face / left-hand / right-hand crops, each crop is embedded through the frozen DINOv3 ViT-B/16 (\S\ref{sec:method:upstream}), and the per-stream CLS sequences are processed by the trained temporal EMA encoders before their output vectors are L2-normalised. We then run $k$-means ($K{=}250$ for the face stream, $K{=}200$ for each hand stream) on these embeddings and select clusters by a farthest-first traversal in centroid space, filtered to clusters whose members span a sufficiently diverse set of source videos (so single-signer / single-shot clusters are excluded). For each chosen cluster we render the 10 nearest-to-centroid crops. The procedure is intended to be \emph{descriptive}: we do not use the cluster labels in any downstream evaluation. Figures~\ref{fig:clusters_face}--\ref{fig:clusters_rh} show the result. Notably, the face stream surfaces clusters defined by \emph{expression and pose} (hand-on-face mouthings, side-profile, neutral closed mouth) that recur across multiple identities --- analogous to \shubert\ cluster~141 --- and each hand stream surfaces \emph{handshape} clusters (pointing / spread-fingers / thumb-up / two-hand contact) that recur across many backgrounds and skin tones, qualitative evidence consistent with sub-lexical grouping, though not proof of signer or background invariance. Anatomical hand labels follow the signer's perspective: ``left hand'' (Figure~\ref{fig:clusters_lh}) corresponds to image-right under our convention, since signers face the camera.

\vspace{0.4em}

\vspace{0.4em}

\vspace{0.4em}

\section{Phonological Feature Recognition}
\label{app:phonology}

\paragraph{Motivation and protocol.}
Following \shubert\ Appendix~B/C, we include a phonological-feature recognition benchmark in addition to task-level evaluation. This probe asks whether a learned video representation preserves the sub-lexical structure from which ASL signs are formed, rather than only supporting a whole-sign class label or an English translation. We adopt this benchmark because it is a direct diagnostic of whether the model has encoded deeper sign-language information: location, handshape, selected fingers, contact, and motion are not incidental visual attributes, but linguistically meaningful components of ASL signs. If a representation supports accurate prediction of these properties across datasets, it provides stronger evidence that the SSL model has captured sign-internal structure and category information.

We use Sem-Lex and ASL Citizen with the ASL-Lex~2.0 feature inventory~\citep{kezar2023semlex,desai2023aslcitizen,sevcikova2021asllex2}. Each masked temporal average feeds 16 linear heads. The \shubert\ rows are published full-fine-tuning results, reported as per-feature top-1 accuracy. For the \ours\ rows, we fully fine-tune the temporal encoders and heads with the image backbone frozen, using Adam with learning rate $10^{-4}$ for both, no weight decay, batch 128, and at most 125 epochs. We use seed 0 for this single-run attribute table, select checkpoints by the mean validation accuracy across supported attributes, and mask missing labels separately for each head's cross-entropy. We preserve the public train/validation/test video splits and the same ASL-Lex label mapping as the comparator. The separately frozen retrieval diagnostic in \S\ref{sec:exp:phonology} does not use these adapted encoders.

\begin{table}[!h]
\centering
\scriptsize
\setlength{\tabcolsep}{3pt}
\begin{tabularx}{\columnwidth}{p{1.9cm}lCC}
\toprule
Phonological feature & Method & \shortstack{Sem-Lex\\R@1} & \shortstack{ASL Citizen\\R@1} \\
\midrule
\multirow{2}{=}{Major Location}        & \shubert       & 84.77 & 90.22 \\
                                       & \textbf{\ours} & \textbf{94.43} & \textbf{94.96} \\
\midrule
\multirow{2}{=}{Minor Location}        & \shubert       & 71.30 & 80.00 \\
                                       & \textbf{\ours} & \textbf{76.83} & \textbf{85.79} \\
\midrule
\multirow{2}{=}{Second Minor Location} & \shubert       & 73.28 & 81.18 \\
                                       & \textbf{\ours} & \textbf{82.34} & \textbf{90.73} \\
\midrule
\multirow{2}{=}{Contact}               & \shubert       & 86.84 & 91.57 \\
                                       & \textbf{\ours} & \textbf{98.93} & \textbf{96.94} \\
\midrule
\multirow{2}{=}{Thumb Contact}         & \shubert       & 84.74 & 87.52 \\
                                       & \textbf{\ours} & \textbf{92.55} & \textbf{92.16} \\
\midrule
\multirow{2}{=}{Sign Type}             & \shubert       & 84.64 & 91.54 \\
                                       & \textbf{\ours} & \textbf{90.72} & \textbf{99.50} \\
\midrule
\multirow{2}{=}{Repeated Movement}     & \shubert       & 82.65 & 89.93 \\
                                       & \textbf{\ours} & \textbf{87.00} & \textbf{96.21} \\
\midrule
\multirow{2}{=}{Path Movement}         & \shubert       & 72.75 & 79.42 \\
                                       & \textbf{\ours} & \textbf{81.12} & \textbf{87.72} \\
\midrule
\multirow{2}{=}{Wrist Twist}           & \shubert       & 90.58 & 93.00 \\
                                       & \textbf{\ours} & \textbf{97.11} & \textbf{99.50} \\
\midrule
\multirow{2}{=}{Selected Fingers}      & \shubert       & 79.53 & 83.44 \\
                                       & \textbf{\ours} & \textbf{89.94} & \textbf{87.67} \\
\midrule
\multirow{2}{=}{Thumb Position}        & \shubert       & 86.04 & 88.19 \\
                                       & \textbf{\ours} & \textbf{97.28} & \textbf{98.76} \\
\midrule
\multirow{2}{=}{Flexion}               & \shubert       & 72.64 & 77.73 \\
                                       & \textbf{\ours} & \textbf{78.74} & \textbf{82.83} \\
\midrule
\multirow{2}{=}{Spread}                & \shubert       & 79.42 & 84.80 \\
                                       & \textbf{\ours} & \textbf{90.99} & \textbf{91.89} \\
\midrule
\multirow{2}{=}{Spread Change}         & \shubert       & 81.60 & 86.58 \\
                                       & \textbf{\ours} & \textbf{86.44} & \textbf{91.75} \\
\midrule
\multirow{2}{=}{Nondominant Handshape} & \shubert       & 76.32 & 84.32 \\
                                       & \textbf{\ours} & \textbf{86.60} & \textbf{93.63} \\
\midrule
\multirow{2}{=}{Handshape}             & \shubert       & 62.93 & 70.80 \\
                                       & \textbf{\ours} & \textbf{71.16} & \textbf{79.51} \\
\midrule
\multirow{2}{=}{Average}               & \shubert       & 79.38 & 85.02 \\
                                       & \textbf{\ours} & \textbf{87.64} & \textbf{91.85} \\
\bottomrule
\end{tabularx}
\caption{Phonological feature recognition on Sem-Lex and ASL Citizen. The table follows the \shubert\ Appendix~B layout; R@1 is top-1 accuracy for each feature-specific classifier. For every feature we report the \shubert\ baseline above and \ours\ below; \ours\ rows are bolded to mark the matched-protocol gain.}
\label{tab:phonology}
\end{table}
\vspace{0.4em}

\paragraph{Phonology results.}
Averaged over the 16 phonological features, \ours\ reaches \emph{87.64}\,R@1 on Sem-Lex and \emph{91.85}\,R@1 on ASL Citizen, compared with \shubert's \emph{79.38} and \emph{85.02}, a matched-protocol gain of \emph{+8.26} and \emph{+6.83} R@1 respectively. \ours\ improves every feature on both datasets, including temporally articulated features --- Path Movement (\emph{+8.37}/\emph{+8.30}), Wrist Twist (\emph{+6.53}/\emph{+6.50}), Repeated Movement (\emph{+4.35}/\emph{+6.28}) --- and fine handshape categories --- Selected Fingers (\emph{+10.41}/\emph{+4.23}), Flexion (\emph{+6.10}/\emph{+5.10}), Handshape (\emph{+8.23}/\emph{+8.71}). On the categorical features that were already strong for \shubert, \ours\ also closes most of the remaining headroom: Major Location reaches \emph{94.43}/\emph{94.96} R@1 (\emph{+9.66}/\emph{+4.74}) and Sign Type reaches \emph{90.72}/\emph{99.50} (\emph{+6.08}/\emph{+7.96}). The consistent gains across the 16 measured attributes indicate that the full \ours\ representation supports more accurate prediction of these sub-lexical categories than the \shubert\ features. This result does not, on its own, distinguish the effects of temporal self-distillation, GLP, and GRT.

To the best of our knowledge, \shubert\ was the first work to report phonological-feature recognition accuracies on ASL Citizen under this setting, and the Sem-Lex numbers are not directly comparable to the original Sem-Lex report because \shubert\ evaluates on the entire public test set. We therefore follow \shubert\ and treat this benchmark as a reusable diagnostic for future sign representation learning work. Here, this benchmark is useful because \ours\ replaces discrete hidden-unit prediction with temporal self-distillation plus geometric supervision; the probe tests whether this continuous temporal objective still preserves the linguistic categories encoded by ASL-Lex. The full feature vocabulary used by the 16 classification heads is listed in Appendix~\ref{app:phonology_details}.

\section{Cross-Sentence t-SNE of Translation-Trained Features}
\label{app:tsne}

\begin{figure*}[!t]
\centering
\includegraphics[width=\textwidth]{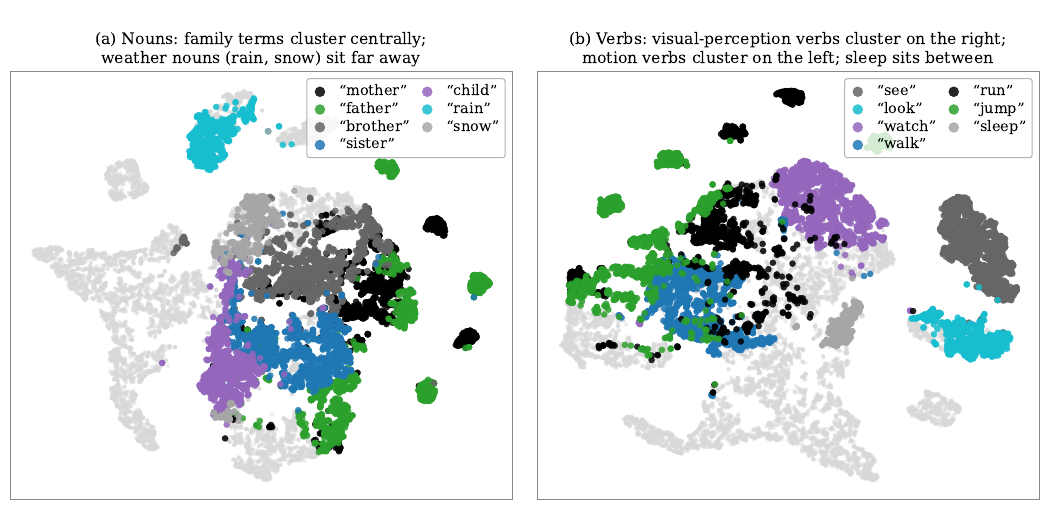}
\caption{t-SNE of \sdeg\ features after SLT training. Noun and verb neighbourhoods are descriptive; this figure is not a geometry diagnostic.}
\label{fig:tsne}
\end{figure*}

Figure~\ref{fig:tsne} shows a t-SNE of \sdeg\ features after SLT training. It visualises post-SLT features from a 2400-word sample spanning How2Sign and the three ISLR datasets, with perplexity 30 and 1000 iterations. It is a descriptive illustration, not a geometric diagnostic and not evidence used in model selection or held-out evaluation.

The noun and verb panels illustrate lexical neighbourhoods after translation supervision. Their projected distances are descriptive; the quantitative geometric evaluation uses the probes in Appendix~\ref{app:geometry_protocol} and signer-separated retrieval in \S\ref{sec:exp:phonology}.

\vspace{0.4em}

\section{Reproducibility Checklist}
\label{app:repro}

This appendix records the training and evaluation protocol. Released artefacts will include per-stream checkpoints, benchmark features, and run configurations. Published baseline values are distinguished from \ours\ task results in Tables~\ref{tab:slt}--\ref{tab:islrfs}; matched geometry controls are reported in Table~\ref{tab:abl_geometry}. Geometry ablation names in this manuscript refer to Eq.~\eqref{eq:total}.

\begin{table}[h]
\centering
\scriptsize
\setlength{\tabcolsep}{3pt}
\begin{tabularx}{\columnwidth}{p{2.45cm}YY}
\toprule
Hyperparameter & First 400K updates & Final 50K updates \\
\midrule
Steps & 400K & 50K \\
Batch (videos / GPU) & 8 & 8 \\
Optimiser & AdamW & AdamW \\
Base LR & $5\!\times\!10^{-4}$ & $2\!\times\!10^{-4}$ \\
LR schedule & 5\% warmup + cosine & 1\% warmup + cosine \\
Weight decay & $0.04 \to 0.4$ & $0.04 \to 0.4$ \\
EMA momentum & $0.994 \to 1.0$ & $0.999 \to 1.0$ \\
Teacher $\tau_T$ & $0.04\to 0.07$ & $0.07$ \\
Student $\tau_S$ & $0.1$ & $0.1$ \\
Multi-temporal crops & $N_g{=}2,\,N_l{=}8$ & same \\
$T_g/T_l$ ranges (contiguous) & $[32,48]/[6,16]$ & same \\
iBOT mask ratio & $25$--$50\%$ & same \\
Loss weights & Eq.~\eqref{eq:total} & Eq.~\eqref{eq:total} \\
Precision & bf16 autocast & bf16 autocast \\
\bottomrule
\end{tabularx}
\caption{\ours\ optimisation settings recorded over the initial 400K and final 50K updates. Both segments use the full loss of Eq.~\eqref{eq:total}.}
\label{tab:pretrain}
\vspace{-0.6em}
\end{table}
\vspace{0.4em}

\begin{table}[h]
\centering
\scriptsize
\setlength{\tabcolsep}{3pt}
\begin{tabularx}{\columnwidth}{p{2.4cm}YY}
\toprule
& Source stage & Target stage \\
\midrule
Data & decontaminated YouTube-ASL & How2Sign / OpenASL train \\
Steps & 250K & 50K \\
Optimiser & AdamW ($\beta_1{=}0.9,\beta_2{=}0.99$) & AdamW \\
LR ByT5 + head & $5\!\times\!10^{-4}$ & $1\!\times\!10^{-4}$ \\
LR \ours\ (live) & $5\!\times\!10^{-5}$ & $1\!\times\!10^{-5}$ \\
Warmup & 10K cosine & 5K cosine \\
Effective batch & 64 utterances & 64 utterances \\
Weight decay & 0.1 & 0.1 \\
Label smoothing & 0.2 & 0.2 \\
Decoding & beam $=5$, $L_{\max}{=}384$ & beam $=5$ \\
Hardware & $4\,\times\,$A100 80GB & $4\,\times\,$A100 80GB \\
\bottomrule
\end{tabularx}
\caption{Translation training hyperparameters. The source stage follows \shubert\ \S4.2; the target stage fine-tunes on the original How2Sign/OpenASL training split and evaluates on the corresponding validation/test splits.}
\label{tab:slt_train}
\vspace{-0.6em}
\end{table}
\vspace{0.4em}

\paragraph{Source split and decontamination.}
For self-supervised pre-training and source-stage SLT training, we follow the \shubert\ public-data protocol. The source corpus is YouTube-ASL~\citep{uthus2023youtubeasl}. Clips that intersect the OpenASL evaluation set~\citep{shi2022openasl} are removed to avoid test contamination, and the removed duration is replaced with non-overlapping ASL videos from YouTube-SL-25~\citep{tanzer_zhang2024ytsl25}. This keeps the public source corpus at approximately 984 source hours. We also do not report MSASL because its test set overlaps with YouTube-ASL, following the same rationale as \shubert.

\paragraph{Data and preprocessing.}
For source pre-training, each YouTube-ASL video is passed through YOLOv8n + ByteTrack to obtain synchronised left-hand, right-hand, and face crop streams. For target-stage fine-tuning, How2Sign and OpenASL videos are processed once from the original training split, matching the main \shubert-style protocol. Validation and test videos are also processed once. Each stream carries a binary validity mask indicating detector failures. Short detector dropouts are filled by ByteTrack's motion model; longer failures are excluded from SSL losses through the validity mask. Crops are square, enlarged $1.4\times$ (hands) and $1.2\times$ (face), and resized to $224\times224$ for the frozen \dinovthree\ encoder. In result tables, pre-training hours are always reported as unique source-video hours, matching \shubert's accounting.

\paragraph{Embedding cache and SSL training.}
Each crop stream is encoded once by the frozen \dinovthree\ ViT-B/16 into a float16 tensor $E^{(r)}\in\mathbb{R}^{T\times768}$. All hand-stream training updates jointly optimise temporal self-distillation, GLP and GRT (Eq.~\eqref{eq:total}); face-stream training uses temporal self-distillation only. Optimisation has an initial block of 400K updates and a final block of 50K updates (Table~\ref{tab:pretrain}), with GLP and GRT active in both. Global and local temporal crops and masked frames follow the settings of Appendix~\ref{app:sdeg_hparams} and Eq.~\eqref{eq:glp}. The frozen hand-model targets are computed for the two hand streams during pre-training and discarded at inference.

\paragraph{Translation.}
The effective batch is $2$ utterances/GPU $\times$ $4$ GPUs $\times$ $8$ accumulation steps $=64$ utterances.
Translation follows \shubert's source-to-target adaptation protocol. The source stage trains the fusion projection and ByT5-Base translation model on the decontaminated YouTube-ASL split for 250K steps. The target stage fine-tunes on the original training split of How2Sign~\citep{duarte2021how2sign} or OpenASL~\citep{shi2022openasl} for 50K steps. FLEURS-ASL~\citep{tanzer2024fleurs_asl} is evaluated zero-shot, since it has no training split in this protocol. Decoding uses beam search with width 5 and maximum output length 384. We report BLEU-4~\citep{papineni2002bleu} computed with SacreBLEU~\citep{post2018sacrebleu} and BLEURT~\citep{sellam2020bleurt} computed with the BLEURT-20 checkpoint, following \shubert.

\paragraph{ISLR, fingerspelling, and phonology.}
ISLR evaluates ASL Citizen, Sem-Lex, and WLASL2000. The stream-fused representation is time-averaged with the attention mask, passed through BatchNorm and a linear classifier, and adapted either as frozen features, rank-1 LoRA, or full fine-tuning as described in \S\ref{sec:exp:islr}. We report Recall@1/5/10 for ASL Citizen and Sem-Lex, and per-instance/per-class top-1 accuracy for WLASL2000. Fingerspelling detection on ASL-STEM-Wiki is trained as per-frame binary classification and evaluated by mean interval-IoU under the dataset cross-validation protocol. The phonological-feature probe in Appendix~\ref{app:phonology} uses 16 simultaneous feature-specific heads and reports Recall@1 for each ASL-Lex feature.

\paragraph{Compute, software, and seeds.}
SSL uses bf16 and AdamW on $4\times$A100 80GB, training streams sequentially: 400K initial updates and 50K final updates, with all losses active in both segments. Table~\ref{tab:compute} reports the corresponding 840 temporal-training GPU-hours and approximately 300 GPU-hours for the appearance cache. WiLoR extraction is separately accounted in Appendix~\ref{app:geometry_protocol}. Software is PyTorch 2.6, CUDA 12.4, Transformers, SacreBLEU, and BLEURT-20. Three-seed summaries are identified explicitly in table captions; single-valued ablations use seed 0. Run configurations will record the seed identifiers and whether pre-training is shared. The matched geometry controls use SSL seed 0, downstream seeds 0/1/2, and validation-only checkpoint selection; Table~\ref{tab:abl_geometry} reports downstream means.

\paragraph{Responsible use.}
Sign-language video contains identifiable signer information. We use the released research corpora under their terms and do not redistribute raw video. Face crops retain non-manual linguistic cues; this is not an anonymisation procedure. Privacy-aware deployment requires a separate data-handling decision informed by prior sign-language research~\citep{rust2024ssvpslt,desai2024_systemic_biases}.

\section{Training and Downstream Diagnostics}
\label{app:diagnostics}

This appendix reports prototype occupancy during SSL and task-specific use of encoder depth. The downstream task and phonological evaluations measure representation utility; the statistics here describe the training distributions.

\subsection{Prototype-use entropy}
\label{app:diag:entropy}

\paragraph{Protocol.}
Let $p_{bt}$ be a teacher frame's Sinkhorn-normalised assignment over $K{=}8192$ prototypes. We report $H(\bar p)/\log K$, where $\bar p$ averages assignments over 1024 held-out frames. This is marginal prototype-use entropy: high values indicate broad occupancy. Balanced Sinkhorn assignments and even identical uniform outputs can have high marginal entropy, so this statistic alone is not a test of feature non-collapse.

\paragraph{Observed trajectory.}
Table~\ref{tab:diag_entropy} shows marginal entropy decreasing from 1.000 at initialisation to about 0.92 for hands and 0.94 for faces. The teacher continues to distribute assignment mass broadly. Differences between streams describe occupancy and do not establish a count of linguistic categories.

\begin{table}[!h]
\centering
\scriptsize
\setlength{\tabcolsep}{4pt}
\begin{tabularx}{\columnwidth}{p{2.7cm}CCC}
\toprule
SSL checkpoint & LH & RH & Face \\
\midrule
init                  & 1.000 & 1.000 & 1.000 \\
100K updates          & 0.962 & 0.958 & 0.971 \\
200K updates          & 0.945 & 0.941 & 0.957 \\
300K updates          & 0.933 & 0.929 & 0.945 \\
400K updates & 0.927 & 0.922 & 0.940 \\
450K updates  & 0.921 & 0.917 & 0.935 \\
\bottomrule
\end{tabularx}
\caption{Marginal prototype-use entropy $H(\bar p)/\log K$ on 1024 held-out frames. Values above 0.91 indicate broad occupancy; balanced assignments alone do not establish non-collapse.}
\label{tab:diag_entropy}
\end{table}
\vspace{0.4em}

\subsection{Layer-weighted-sum head weights}
\label{app:diag:layerweights}

\paragraph{Protocol.}
For each downstream task and each stream, the fusion head learns a softmax distribution over the $L{+}1{=}7$ \ours encoder layers ($L{=}0$: projected \dinovthree\ input; $L{=}1{..}6$: transformer blocks), as described in \S\ref{sec:exp:slt} and \S\ref{sec:exp:islr}. We report the converged softmax weights after the target-stage / downstream training completes, averaged over three random seeds. Each row of Table~\ref{tab:diag_layerweights} is therefore a probability distribution and sums to $1.0$.

\paragraph{Observed patterns.}
The hand-stream weights peak at block 5 (0.21--0.27); recognition and fingerspelling place more combined weight on blocks 5 and 6 than translation does. Face weights are flatter. Because mixing coefficients depend on feature scale and are learned during downstream training, they describe how the head combines layers rather than proving a uniquely localised linguistic mechanism.

\begin{table}[!h]
\centering
\scriptsize
\setlength{\tabcolsep}{2pt}
\begin{tabularx}{\columnwidth}{p{1.5cm}p{0.7cm}CCCCCCC}
\toprule
Task & Stream & $L{=}0$ & $L{=}1$ & $L{=}2$ & $L{=}3$ & $L{=}4$ & $L{=}5$ & $L{=}6$ \\
\midrule
\multirow{3}{*}{SLT (H2S)}   & LH   & 0.06 & 0.09 & 0.12 & 0.16 & 0.18 & 0.21 & 0.18 \\
                             & RH   & 0.05 & 0.08 & 0.11 & 0.15 & 0.19 & 0.22 & 0.20 \\
                             & Face & 0.07 & 0.11 & 0.14 & 0.17 & 0.19 & 0.18 & 0.14 \\
\midrule
\multirow{3}{*}{ISLR (ASLC)} & LH   & 0.04 & 0.07 & 0.10 & 0.14 & 0.18 & 0.24 & 0.23 \\
                             & RH   & 0.04 & 0.07 & 0.10 & 0.14 & 0.18 & 0.24 & 0.23 \\
                             & Face & 0.09 & 0.12 & 0.15 & 0.17 & 0.17 & 0.16 & 0.14 \\
\midrule
\multirow{3}{*}{FS (StemWiki)} & LH   & 0.03 & 0.05 & 0.08 & 0.13 & 0.20 & 0.27 & 0.24 \\
                               & RH   & 0.03 & 0.05 & 0.08 & 0.13 & 0.20 & 0.27 & 0.24 \\
                               & Face & 0.11 & 0.14 & 0.16 & 0.17 & 0.16 & 0.14 & 0.12 \\
\bottomrule
\end{tabularx}
\caption{Converged layer-weighted-sum softmax weights per task and per stream. Each row is a probability distribution over the seven encoder layers ($L{=}0$: projected \dinovthree\ input, $L{=}1{..}6$: transformer blocks) and sums to $1.0$. Values are averaged over three random seeds of the target-stage / downstream training.}
\label{tab:diag_layerweights}
\end{table}
\vspace{0.4em}

\subsection{Pre-Training Compute Budget}
\label{app:compute}

Table~\ref{tab:compute} reports the wall-clock and GPU-hour budget of the \ours\ SSL pipeline on the $4\,\times\,$A100 80\,GB hardware used for every main run in this paper. The three per-stream temporal Transformers are trained sequentially: each stream consumes the full 4-GPU pool for the duration of its optimisation schedule, then the next stream begins. With the frozen-\dinovthree\ embedding cache enabled (\S\ref{sec:exp:setup}), per-step throughput is dominated by the temporal Transformer rather than by the image-backbone forward pass, which is what allows the full upstream pipeline to fit inside this budget. The embedding-cache row records the one-off preprocessing cost referenced in \S\ref{sec:exp:setup} and is amortised across all SSL ablations and downstream tasks. The final row reports the upstream SSL cost quoted by \shubert~\citep{gueuwou2025shubert} as a reference point.

\begin{table}[!h]
\centering
\scriptsize
\setlength{\tabcolsep}{4pt}
\begin{tabularx}{\columnwidth}{p{3.2cm}CCC}
\toprule
Stage & Per-stream GPU-h & Per-stream wall-clock & Total GPU-h \\
\midrule
First 400K updates              & 240 & ${\approx}60$ h  & 720 \\
Final 50K updates  & 40  & ${\approx}10$ h   & 120 \\
\midrule
SSL subtotal (3 streams)          & 280 & ${\approx}70$ h  & 840 \\
Embedding cache (one-off)         & --- & ---              & ${\approx}300$ \\
\midrule
\shubert\ upstream (ref.)         & --- & ---              & 1344 \\
\bottomrule
\end{tabularx}
\caption{\ours\ pre-training compute budget on $4\,\times\,$A100 80\,GB. Streams are trained sequentially, so the wall-clock for the full three-stream upstream is approximately $3{\times}$ the per-stream wall-clock plus negligible scheduling overhead. The embedding-cache row is the one-off preprocessing cost discussed in \S\ref{sec:exp:setup}. The \shubert\ row reports the upstream-only figure of $1344$~GPU-hours quoted by~\citet{gueuwou2025shubert} ($8\,\times\,$A6000 $\times\,7$~days).}
\label{tab:compute}
\end{table}
\vspace{0.4em}

\subsection{SSL Data-Scaling Curve}
\label{app:datascale}

Table~\ref{tab:abl_scale} reports the two-point endpoints of the SSL data-scaling scan at $10\%$ and $100\%$ of the decontaminated YouTube-ASL source corpus. Table~\ref{tab:datascale} extends this scan to a five-point pre-training-hours sweep along the same corpus, evaluated under the same frozen-features setup as Table~\ref{tab:abl_scale} (\ours encoders frozen and the same downstream ByT5-Base head). For every row, the temporal sampler (Table~\ref{tab:pretrain}), embedding cache, downstream source-stage + target-stage recipe (Table~\ref{tab:slt_train}), and SSL optimisation schedule are held fixed at the defaults; only the size of the source subset used for SSL pre-training differs. Intermediate subsets are drawn as a video-disjoint random sample of the full source-video pool to keep the same speaker and topic distribution at every scale.

\begin{table}[!h]
\centering
\scriptsize
\setlength{\tabcolsep}{3pt}
\begin{tabularx}{\columnwidth}{p{2.7cm}CCC}
\toprule
Pre-train hours & BLEU-1 & BLEU & BLEURT \\
\midrule
98 (10\%) & 33.6 & 13.5 & 46.1 \\
984 (100\%) & 37.9 & 16.8 & 50.4 \\
\bottomrule
\end{tabularx}
\caption{Pre-training data scale (\shubert\ Tab.~5 analogue), evaluated on the \emph{How2Sign test} split with the \ours encoders frozen and the same downstream head as Table~\ref{tab:abl_mask}. Two-point scan on 10\% vs 100\% of the decontaminated YouTube-ASL source corpus; PT hours follow the unique source-corpus accounting used in Table~\ref{tab:slt}. Table~\ref{tab:datascale} extends this to a five-point sweep.}
\label{tab:abl_scale}
\vspace{-0.6em}
\end{table}
\vspace{0.4em}

\begin{table}[!h]
\centering
\scriptsize
\setlength{\tabcolsep}{4pt}
\begin{tabularx}{\columnwidth}{p{2.6cm}CC}
\toprule
Pre-train hours & H2S BLEU & H2S BLEURT \\
\midrule
98  \,\,(10\%)   & 13.5 & 46.1 \\
246 \,\,(25\%)   & 14.8 & 47.9 \\
492 \,\,(50\%)   & 16.0 & 49.4 \\
738 \,\,(75\%)   & 16.5 & 50.0 \\
984 \,\,(100\%)  & 16.8 & 50.4 \\
\bottomrule
\end{tabularx}
\caption{\ours\ SSL data-scaling curve. Frozen-features H2S BLEU and BLEURT as a function of the source-corpus pre-training budget. The $10\%$ and $100\%$ rows reproduce the corresponding rows of Table~\ref{tab:abl_scale}. The curve is monotonic and not yet flat at $100\%$ of the public corpus, consistent with the data-scaling regime reported by \shubert~\citep{gueuwou2025shubert}; increasing hours from 98 to 984 yields 3.3 BLEU, with smaller gains at the upper end.}
\label{tab:datascale}
\end{table}
\vspace{0.4em}

\subsection{Additional Ablations}
\label{app:moreabl}

This appendix gives the full tables for the design-axis and \shubert-protocol ablations summarised in \S\ref{sec:exp:ablations}: masking strategy (Table~\ref{tab:abl_mask}), layer pooling (Table~\ref{tab:abl_layer}), frozen vs.\ live fine-tuning (Table~\ref{tab:abl_ft}), teacher crop-length randomisation (Table~\ref{tab:abl_windowsize}), visual backbone (Table~\ref{tab:abl_backbone}), translation decoder (Table~\ref{tab:abl_decoder}), and hand detector and hand geometry model (Table~\ref{tab:abl_handmodel}). The two-point and five-point data-scale results are in Tables~\ref{tab:abl_scale}--\ref{tab:datascale} above. All rows reuse the common training protocol of Appendix~\ref{app:repro} and are evaluated on the \emph{How2Sign test} split unless noted; BLEU denotes corpus BLEU-4.

\begin{table}[!h]
\centering
\scriptsize
\setlength{\tabcolsep}{3pt}
\begin{tabularx}{\columnwidth}{p{2.7cm}CCC}
\toprule
Strategy & BLEU-1 & BLEU & BLEURT \\
\midrule
Channel masking & 35.2 & 14.4 & 47.6 \\
Time masking & 36.1 & 14.9 & 48.1 \\
Random span masking & 37.0 & 15.7 & 48.9 \\
\midrule
Frame mask only (\ours) & 37.9 & 16.8 & 50.4 \\
Frame + stream mask & 37.3 & 16.3 & 49.7 \\
\bottomrule
\end{tabularx}
\caption{Masking-pattern controls within \ours, with frozen encoders and the same downstream head. Pattern names follow \shubert's masking comparison; the cells are \ours\ results, not published \shubert\ values. Frame-only masking gives the highest BLEURT.}
\label{tab:abl_mask}
\vspace{-0.6em}
\end{table}
\vspace{0.4em}

\paragraph{Layer pooling.}
Table~\ref{tab:abl_layer} confirms that the \ours temporal encoder makes a substantial contribution to translation: feeding the raw frozen \dinovthree\ per-frame embeddings to ByT5 without the \ours encoders drops to \emph{9.4}~BLEU/\emph{41.2}~BLEURT, a \emph{-7.4}~BLEU gap from the weighted-sum default. The last-layer-only variant (\emph{14.7}~BLEU/\emph{47.8}~BLEURT) recovers most but not all of the gap, motivating the learnable softmax over all $L{+}1$ encoder layers as the default pooling strategy.

\begin{table}[!h]
\centering
\scriptsize
\setlength{\tabcolsep}{3pt}
\begin{tabularx}{\columnwidth}{p{2.6cm}CCC}
\toprule
Layer of \ours\ & BLEU-1 & BLEU & BLEURT \\
\midrule
None (raw \dinovthree) & 24.3 & 9.4 & 41.2 \\
Last layer & 34.6 & 14.7 & 47.8 \\
Weighted sum (default) & 37.9 & 16.8 & 50.4 \\
\bottomrule
\end{tabularx}
\caption{Layer pooling (\shubert\ Tab.~6 analogue), with the \ours encoders frozen and the same downstream head as Table~\ref{tab:abl_mask}. The ``None'' row feeds the raw \dinovthree\ per-frame embeddings to ByT5 without the \ours encoders.}
\label{tab:abl_layer}
\vspace{-0.6em}
\end{table}
\vspace{0.4em}

\paragraph{Frozen vs.\ live fine-tune.}
The two rows of Table~\ref{tab:abl_ft} reproduce the gap reported in the main Table~\ref{tab:slt}: \emph{16.8}~BLEU frozen vs.\ \emph{17.9}~BLEU live (BLEURT \emph{50.4}$\to$\emph{51.3}, BLEU-1 \emph{37.9}$\to$\emph{39.4}). The \emph{+1.1}~BLEU gain from live fine-tuning is small relative to the cost of updating the temporal encoder; the frozen variant is therefore the recommended operating point when downstream compute is constrained, and live fine-tuning is the recommended operating point for the highest score in this comparison.

\begin{table}[!h]
\centering
\scriptsize
\setlength{\tabcolsep}{3pt}
\begin{tabularx}{\columnwidth}{p{2.6cm}CCC}
\toprule
Fine-tune \ours? & BLEU-1 & BLEU & BLEURT \\
\midrule
$\times$ frozen & 37.9 & 16.8 & 50.4 \\
\checkmark\ live fine-tune & 39.4 & 17.9 & 51.3 \\
\bottomrule
\end{tabularx}
\caption{Frozen vs.\ fine-tune (\shubert\ Tab.~7 analogue). This isolates whether downstream supervision benefits from updating the temporal \ours encoder beyond the frozen-feature setting; the two rows reproduce the corresponding ``frozen'' and ``live fine-tune'' rows of Table~\ref{tab:slt}.}
\label{tab:abl_ft}
\vspace{-0.6em}
\end{table}
\vspace{0.4em}

\paragraph{Teacher crop-length randomisation.}
The current default sampler draws global crops from $\mathcal{U}\{32,\ldots,48\}$ and local crops from $\mathcal{U}\{6,\ldots,16\}$, independently on each update (\S\ref{sec:method:ssl}). At $N_g{=}2$ and $N_l{=}8$, this amounts to an expected $2\cdot40+8\cdot11=168$ sampled frames per update. Table~\ref{tab:abl_windowsize} compares three fixed-window configurations $(T_l,T_g)\in\{(16,64),(20,80),(24,72)\}$, which use respectively 256, 320, and 336 sampled frames per update. The recorded BLEU-4 values are 17.3, 17.5, and 17.4, versus 17.9 for the randomised default. Because the frame budgets differ, the comparison is descriptive.

\begin{table}[!h]
\centering
\scriptsize
\setlength{\tabcolsep}{4pt}
\begin{tabularx}{\columnwidth}{p{4.0cm}CC}
\toprule
$T_l\,/\,T_g$ sampler & H2S BLEU & $\Delta$ \\
\midrule
Fixed $16\,/\,64$                      & $17.3_{\pm 0.15}$ & $-0.6$ \\
Fixed $20\,/\,80$                      & $17.5_{\pm 0.15}$ & $-0.4$ \\
Fixed $24\,/\,72$                      & $17.4_{\pm 0.15}$ & $-0.5$ \\
\midrule
Random $\mathcal{U}\{6{-}16\}\,/\,\mathcal{U}\{32{-}48\}$ (default) & $\mathbf{17.9_{\pm 0.15}}$ & --- \\
\bottomrule
\end{tabularx}
\caption{Crop-window comparison. The default is $T_l\in[6,16]$, $T_g\in[32,48]$; the fixed-window rows use longer windows and larger frame budgets. The last row reproduces the live How2Sign entry of Table~\ref{tab:slt}.}
\label{tab:abl_windowsize}
\vspace{-0.6em}
\end{table}
\vspace{0.4em}

\paragraph{Visual backbone.}
The frozen frame backbone is varied while the temporal encoder, fusion module, decoder, and training schedule are fixed (Table~\ref{tab:abl_backbone}). H2S BLEU ranges from 16.5 to 18.0 across five DINO backbones. \dinovthree\ ViT-B/16 reaches 17.9, while ViT-L/16 adds 0.1 BLEU at approximately 3.5 times the image-backbone parameters. These results support the B/16 default and show that the complete pipeline works across backbone sizes; geometry-specific effects are isolated by Table~\ref{tab:abl_geometry}.

\begin{table}[!h]
\centering
\scriptsize
\setlength{\tabcolsep}{3pt}
\begin{tabularx}{\columnwidth}{p{2.7cm}CCC}
\toprule
Frozen per-frame backbone & Params & H2S BLEU & BLEURT \\
\midrule
\dinov\ ViT-S/14~\citep{oquab2023dinov2}            & 21M  & 16.5 & 49.4 \\
\dinov\ ViT-L/14~\citep{oquab2023dinov2}            & 300M & 17.4 & 50.5 \\
\dinovthree\ ViT-S/16~\citep{simeoni2025dinov3}     & 22M  & 17.2 & 50.1 \\
\dinovthree\ ViT-B/16 (default)~\citep{simeoni2025dinov3} & 86M  & 17.9 & 51.3 \\
\dinovthree\ ViT-L/16~\citep{simeoni2025dinov3}     & 304M & 18.0 & 51.4 \\
\bottomrule
\end{tabularx}
\caption{Visual backbone ablation with \ours\ live fine-tuned (same regime as the \ours\ live row of Table~\ref{tab:slt}). The backbone is always frozen; only the per-frame CLS token is propagated to the \ours temporal encoder. SSL pre-training is re-run per row; downstream translation training (Table~\ref{tab:slt_train}) is identical across rows.}
\label{tab:abl_backbone}
\vspace{-0.6em}
\end{table}
\vspace{0.4em}

\paragraph{Translation decoder.}
Table~\ref{tab:abl_decoder} varies the decoder at the live fine-tuning operating point. Both upstreams use the source-to-target schedule of Table~\ref{tab:slt_train}; decoder settings follow their releases. \ours\ exceeds the \shubert\ comparison by 1.4--1.7 BLEU across these five decoders. ByT5-Base achieves the highest score in this comparison: ByT5-Large loses 0.4 BLEU for \ours, and mT5-Base/mBART-large trail by 0.8/1.1. This tests consistency across the evaluated decoders, rather than a decoder-independent ordering for all architectures.

\begin{table}[!h]
\centering
\scriptsize
\setlength{\tabcolsep}{3pt}
\begin{tabularx}{\columnwidth}{p{2.5cm}CCC}
\toprule
Translation decoder & Params & \shubert\ BLEU & \ours\ BLEU \\
\midrule
ByT5-Small~\citep{xue2022byt5}            & 300M  & 15.0 & 16.4 \\
ByT5-Base (default)~\citep{xue2022byt5}   & 582M  & 16.2 & 17.9 \\
ByT5-Large~\citep{xue2022byt5}            & 1.23B & 15.9 & 17.5 \\
mT5-Base~\citep{xue2021mt5}               & 580M  & 15.5 & 17.1 \\
mBART-large~\citep{liu2020mbart}          & 610M  & 15.1 & 16.8 \\
\bottomrule
\end{tabularx}
\caption{Translation decoder ablation with both upstreams (\shubert\ and \ours) at live fine-tune. All rows use the same source-stage + target-stage recipe (Table~\ref{tab:slt_train}); only the decoder backbone and its $d_{\mathrm{model}}$ projection change. The ByT5-Base default cells reproduce the corresponding \shubert\ (\emph{16.2}) and \ours\ live (\emph{17.9}) rows of Table~\ref{tab:slt}.}
\label{tab:abl_decoder}
\vspace{-0.6em}
\end{table}
\vspace{0.4em}

\paragraph{Hand detector and hand geometry model.}
\label{app:handmodel}
Table~\ref{tab:abl_handmodel} varies the two hand-specific components of the pipeline under the matched-control protocol of Table~\ref{tab:abl_geometry}. The upper block replaces the geometry model that supplies the GLP and GRT targets, with crops from YOLOv8n + ByteTrack. HaMeR~\citep{pavlakos2024hamer} has no MANO pose-query tokens, so it is compared with a WiLoR variant that also drops the latent target $u_t$ and keeps $\theta_t$ and $J_t$. The lower block replaces the hand detector that produces the crops for both \dinovthree\ and the geometry model, with WiLoR targets.

\begin{table}[!h]
\centering
\scriptsize
\setlength{\tabcolsep}{3pt}
\begin{tabularx}{\columnwidth}{p{4.1cm}CC}
\toprule
Variant & H2S BLEU & ASLC R@1 \\
\midrule
\multicolumn{3}{l}{\emph{Geometry model} (detector: YOLOv8n + ByteTrack)} \\
None (SD only) & 15.6 & 0.642 \\
WiLoR, $u_t,\theta_t,J_t$ (default) & 17.9 & 0.704 \\
WiLoR, $\theta_t,J_t$ (no $u_t$) & 16.8 & 0.670 \\
HaMeR, $\theta_t,J_t$ & 17.6 & 0.705 \\
\midrule
\multicolumn{3}{l}{\emph{Hand detector} (geometry: WiLoR)} \\
YOLOv8n + ByteTrack (default) & 17.9 & 0.704 \\
YOLOv8n only (no ByteTrack) & 16.4 & 0.61 \\
WiLoR detector + ByteTrack & 17.5 & 0.694 \\
MediaPipe Hands + ByteTrack & 18.0 & 0.697 \\
\bottomrule
\end{tabularx}
\caption{Hand geometry model and hand detector substitutions under the matched-control protocol of Table~\ref{tab:abl_geometry} (984-hour source split, 450K updates, identical downstream schedules; H2S uses live SLT, ASLC full temporal fine-tuning). The default, SD-only, and no-ByteTrack rows reproduce Tables~\ref{tab:abl_geometry} and~\ref{tab:abl_ours}.}
\label{tab:abl_handmodel}
\vspace{-0.6em}
\end{table}
\vspace{0.4em}

\section{Ethics Statement}
\label{app:ethics}

We use publicly released ASL corpora under their research terms and do not redistribute raw videos: YouTube-ASL and non-overlapping replacement clips from YouTube-SL-25 for source training; How2Sign, OpenASL, and FLEURS-ASL for translation; ASL Citizen, Sem-Lex, and WLASL2000 for recognition; and ASL-STEM-Wiki for fingerspelling. The frozen WiLoR model imports supervised hand-reconstruction knowledge. Aggregate benchmark scores do not establish equal performance across signers, skin tones, or signing styles. Face crops preserve linguistic non-manual signals and remain identifiable. Any deployment should address consent, privacy, and failure modes with Deaf-community participation~\citep{desai2024_systemic_biases}.

\section{Sample How2Sign Translations}
\label{app:translations}

Table~\ref{tab:appendix_h2s} retains the supplied How2Sign translation examples alongside published \shubert\ hypotheses. Rows 1--6 show complete or shortened references; row 7 groups short fragments and is illustrative, not one aligned utterance. These selected examples illustrate translation differences and do not define a quantitative evaluation set.

\begin{table}[!h]
\centering
\scriptsize
\setlength{\tabcolsep}{2.5pt}
\begin{tabularx}{\columnwidth}{p{0.30cm}p{1.2cm}Y}
\toprule
\# & Source & Sentence \\
\midrule
(1) & Reference & \emph{And that's a great vital point technique for women's self defense.} \\
    & \shubert & \emph{This is a really great point for self defense.} (verbatim from \citet{gueuwou2025shubert} Tab.~11) \\
    & \ours & \emph{And that's a great vital technique for women's self defense.} \\
\midrule
(2) & Reference & \emph{In this clip I'm going to show you how to tape your cables down.} \\
    & \shubert & \emph{In this clip I'm going to show you how to brand out the cable strings.} \\
    & \ours & \emph{In this clip I'm going to show you how to tape down your cables.} \\
\midrule
(3) & Reference & \emph{In this segment we're going to talk about how to load your still for distillation of lavender essential oil.} \\
    & \shubert & \emph{In this clip we're going to talk about how to take our stick for disinfectant oil.} \\
    & \ours & \emph{In this clip we're going to talk about how to load your still for the distillation of lavender oil.} \\
\midrule
(4) & Reference & \emph{You are dancing, and now you are going to need the veil and you are going to just grab the veil as far as possible.} \\
    & \shubert & \emph{Her dancing and now now she needs her feather to grab it with her foot as far as possible.} \\
    & \ours & \emph{You are dancing and now you need the veil, and you're going to grab the veil as far as possible.} \\
\midrule
(5) & Reference & \emph{But if you have to setup a new campfire, there's two ways to do it in a very low impact; one is with a mound fire ...} \\
    & \shubert & \emph{But if you have to set a new campfire, there are two ways to do a low impact one ...} \\
    & \ours & \emph{But if you have to set up a new campfire, there are two low-impact ways to do it; one is with a mound fire.} \\
\midrule
(6) & Reference & \emph{So, this is a very important part of the process.} \\
    & \shubert & \emph{This is a very important part of the process.} \\
    & \ours & \emph{So, this is a very important part of the process.} \\
\midrule
(7) & Reference & \emph{thank you / come on / now I've come this far ...} (representative short How2Sign segment) \\
    & \shubert & \emph{thank you / come on / how do you feel about it} \\
    & \ours & \emph{thank you / come on / now I have come this far} \\
\bottomrule
\end{tabularx}
\caption{Selected qualitative How2Sign examples; row 7 contains grouped fragments. \shubert\ hypotheses are verbatim from~\citet{gueuwou2025shubert} Tables 11 and 13 (deduplicated and shortened).}
\label{tab:appendix_h2s}
\end{table}
\vspace{0.4em}

\section{ASL Phonological Feature Classification Details}
\label{app:phonology_details}

Like spoken languages, American Sign Language can be described through a system of phonological features. These features capture essential components of sign formation, including hand configuration, movement pattern, body-relative location, and contact. Following \shubert\ Appendix~C, we use the ASL-Lex~2.0 feature set and list below the label values used for the 16 classification heads. A few feature values appear in only one of Sem-Lex or ASL Citizen; we keep the same feature vocabulary as \shubert\ to preserve benchmark comparability.

\paragraph{Handshape.}
v, 5, y, h, open\_b, c, baby\_o, flat\_h, o, l, 1, a, open\_8, w, curved\_5, d, flatspread\_5, i, f, s, p, flat\_b, curved\_4, flat\_o, g, open\_e, 4, closed\_b, bent\_1, 3, flat\_horns, goody\_goody, flat\_m, bent\_v, flat\_1, r, 8, curved\_v, open\_h, curved\_1, horns, flat\_ily, flat\_n, bent\_l, stacked\_5, ily, e, flat\_v, curved\_l, spread\_open\_e, curved\_h, 7, closed\_e, t, flat\_4, open\_f, k, and spread\_e.

\paragraph{Nondominant Handshape.}
v, 5, y, none, open\_b, Dominance Condition Violation, B, 1, a, open\_8, C, s, h, o, flat\_b, curved\_5, p, c, S, closed\_b, 4, flat\_m, bent\_v, flat\_1, flat\_h, baby\_o, curved\_v, i, f, bent\_1, Symmetry Violation, flatspread\_5, flat\_o, curved\_1, open\_h, stacked\_5, g, l, bent\_l, 3, 8, spread\_open\_e, e, horns, w, r, Lax, curved\_l, open\_e, flat\_4, O, curved\_b, A, ily, flat\_v, and flat\_horns.

\paragraph{Minor Location.}
Neutral, Head Away, Body Away, Hand Away, Palm, Finger Tip, Forehead, Finger Front, Mouth, Chin, Other, Upper Arm, Torso Top, Forearm Back, Cheek Nose, Wrist Front, Palm Back, Finger Back, Finger Radial, Under Chin, Finger Ulnar, Wrist Back, Shoulder, Arm Away, Forearm Ulnar, Torso Mid, Heel, Clavicle, Eye, Forearm Front, Neck, Torso Bottom, Upper Lip, Head Top, Elbow Back, Hips, and Waist.

\paragraph{Second Minor Location.}
Neutral, Head Away, Torso Bottom, Finger Tip, Hand Away, none, Palm, Forearm Back, Finger Back, Body Away, Torso Top, Finger Front, Chin, Arm Away, Upper Arm, Finger Ulnar, Eye, Hips, Neck, Palm Back, Forearm Front, Finger Radial, Mouth, Heel, Torso Mid, Other, Waist, Cheek Nose, Forehead, Elbow Back, Under Chin, Clavicle, Shoulder, Forearm Ulnar, Head Top, Upper Lip, and Forearm Radial.

\paragraph{Sign Type.}
Symmetrical Or Alternating, One Handed, Dominance Violation, Asymmetrical Different Handshape, Asymmetrical Same Handshape, and Symmetry Violation.

\paragraph{Path Movement.}
Curved, Back And Forth, Straight, Circular, None, Z-shaped, Other, and X-shaped.

\paragraph{Flexion.}
Fully Open, Curved, Bent, Flat, none, Fully Closed, Stacked, and Crossed.

\paragraph{Selected Fingers.}
im, imrp, p, i, t, m, ip, imp, mr, imr, r, and mrp.

\paragraph{Major Location.}
Neutral, Head, Body, Hand, and Arm.

\paragraph{Spread Change.}
$1.0$, $0.0$, and none.

\paragraph{Thumb Contact.}
$1.0$, $0.0$, and none.

\paragraph{Spread.}
$1.0$, $0.0$, and none.

\paragraph{Thumb Position.}
Closed and Open.

\paragraph{Repeated Movement.}
$1.0$ and $0.0$.

\paragraph{Contact.}
$1.0$ and $0.0$.

\paragraph{Wrist Twist.}
$0.0$ and $1.0$.

\paragraph{Interpretation.}
These classifiers decompose performance into linguistically meaningful categories. Location and contact assess spatial cues; handshape and finger features assess manual articulation; path movement, repetition, and wrist twist assess temporal properties. They evaluate labels independent of WiLoR. Attribution to GLP and GRT is tested by the matched objectives and cross-signer retrieval rather than by comparing only to an external SSL system.

\section{Discussion}
\label{app:discussion}

\paragraph{Why geometry enters an intermediate block.}
The temporal appearance pathway provides contextual frame features. GLP at block 4 supervises those features with an articulated target while leaving two subsequent blocks for the sequence and relational objectives. The weighted downstream sum can use both intermediate and final states. The block-6 control in Table~\ref{tab:abl_geometry} tests whether supervising only the final representation loses some of this benefit. The design specifies an interface for geometry within temporal learning, rather than assuming that encoder depth automatically corresponds to a linguistic hierarchy.

\paragraph{Relationship to \shubert\ and \signdinobaseline.}
\shubert\ learns discrete hidden-unit targets across signing streams, and the CVPR \signdinobaseline~\citep{gan2026signdino} learns from frame-level articulator-focused views. \ours\ uses continuous temporal self-distillation and explicit hand-model targets. Published results contextualise task quality; shared evaluation heads do not make different upstreams identical experiments. The matched controls in Table~\ref{tab:abl_geometry} hold the appearance backbone and schedule fixed, directly testing geometry's contribution within \ours.

\paragraph{Interpreting the smaller live fine-tuning gain.}
Frozen and live \ours\ differ by 1.1 BLEU on How2Sign (16.8 to 17.9). Frozen features are useful without updating the temporal encoder; live fine-tuning improves task quality further. This within-model comparison isolates downstream adaptation and does not compare adaptation gains from different published ablation protocols.

\end{document}